\documentclass[letterpaper]{article}
\usepackage[preprint]{aaai2027}
\usepackage[hyphens]{url}
\usepackage{graphicx}
\usepackage{natbib}
\usepackage{caption}

\usepackage[hyphens]{url}  
\usepackage{graphicx} 
\usepackage{natbib}  
\usepackage{caption} 
\usepackage{algorithm}
\usepackage{algorithmic}

\usepackage{newfloat}
\usepackage{amsmath}
\usepackage{amssymb}
\usepackage{amsthm}
\usepackage{dsfont}

\usepackage{booktabs}
\usepackage{makecell}
\usepackage{xcolor}

\newtheorem{theorem}{Theorem}

\newtheorem{lemma}{Lemma}
\newtheorem{corollary}{Corollary}
\newtheorem{assumption}{Assumption}
\newtheorem{definition}{Definition}
\theoremstyle{remark}
\newtheorem{remark}{Remark}
\theoremstyle{plain}

\newcommand{\E}{\mathbb{E}}
\newcommand{\Prob}{\mathbb{P}}
\newcommand{\ind}{\mathds{1}}
\newcommand{\esssup}{\operatorname*{ess\,sup}}
\newcommand{\essinf}{\operatorname*{ess\,inf}}
\newcommand{\supnorm}[1]{\left\lVert #1 \right\rVert_\infty}

\newcommand{\Zn}[1]{Z_{n}^{#1}}
\newcommand{\Top}{\mathcal{T}^{\,n}_{\tau}}
\newcommand{\Qfix}{Q^{n}_{\tau}}
\newcommand{\kt}{\kappa_\tau}
\newcommand{\Vinf}{V^{\infty}}
\newcommand{\sg}{\operatorname{sg}}

\newcommand{\barnum}[1]{$\overline{\text{#1}}$}

\DeclareCaptionStyle{ruled}{labelfont=normalfont,labelsep=colon,strut=off} 

\title{Upper-Expectile Multi-Step Q-Learning for Off-Policy Reinforcement Learning}

\author{
  Abdelghani Ghanem \quad\quad Mounir Ghogho
}

\affiliations{
  College of Computing\\
  Mohammed VI Polytechnic University\\
  Rabat 11103, Morocco\\
  \texttt{\{abdelghani.ghanem-ext,mounir.ghogho\}@um6p.ma}
}

\begin{document}

\maketitle

\begin{abstract}
Multi-step returns accelerate reward propagation in off-policy
reinforcement learning, but couple the evaluation of each decision to
the suboptimal logged actions that follow it, inducing a pessimistic
bias that grows with the horizon. We propose Expectile $n$-step
Q-learning (ENQ), which replaces the symmetric $n$-step
temporal-difference (TD) loss with an asymmetric expectile loss on the
action-value error, with expectile level $\tau$ as the only
method-specific hyperparameter added beyond $n$-step TD. We prove that
the ENQ operator is a $\gamma^{n}$-contraction. Under deterministic
dynamics, at $\tau=1$, its bias vanishes at the optimal action-value
function $Q^*$ on covered in-support pairs, and the corresponding fixed point satisfies the separation-$n$ instance
and its multiples of the lower-bound inequality used by Long-Horizon
Q-learning (LQL).
Under stochastic dynamics, the operator bias admits two-sided bounds
with horizon-independent noise constants. Using a single expectile
level $\tau=0.8$ and a fixed backup horizon across 27 manipulation and
navigation task instances, ENQ is competitive with LQL on aggregate,
achieves higher measured training-step throughput in our profiling
study, and benefits more from a ten-critic ensemble in a controlled
scaling experiment.
\end{abstract}

\section{Introduction}

\begin{figure}[t] 
    \centering
    \includegraphics[width=\columnwidth]{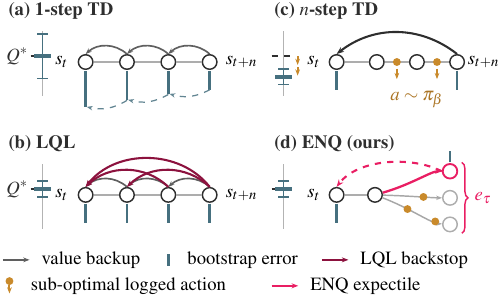} 
\caption{\textbf{Bootstrap error and behavioral bias in long-horizon value learning.}
\textbf{(a)} One-step TD compounds bootstrap errors across successive backups.
\textbf{(b)} LQL limits this propagation using trajectory-level constraints.
\textbf{(c)} $n$-step TD removes intermediate bootstrapping but inherits pessimistic bias from suboptimal logged actions.
\textbf{(d)} ENQ fits an upper expectile of logged $n$-step targets, reducing the influence of poor continuations while retaining multi-step reward propagation.}

    \label{fig:intro_figure}
\end{figure}

Off-policy value-based reinforcement learning (RL) aims to learn improved policies from logged experience collected by prior policies, human operators, or exploratory agents, by training a Q-function to agree with the immediate reward plus a discounted estimate of the counterfactual continuation under the policy being learned \citep{watkins1992q,sutton1998reinforcement,levine2020offline}. In long-horizon continuous control and robotics, where interaction is expensive and reward sparse, using logged transitions efficiently is essential, yet bootstrapping over long horizons is brittle: estimation errors at later states propagate backward through temporal-difference (TD) updates and compound \citep{van2018deep,fujimoto2019off,kumar2019stabilizing}.

The classical remedy, the $n$-step return, replaces the $1$-step backup with a segment of $n$ observed rewards before bootstrapping, shortening the effective bootstrap chain by a factor of $n$ \citep{sutton1988learning,hessel2018rainbow,fedus2020revisiting}. Off-policy, this remedy carries a well-known cost: the $n$-step target for $Q(s_t,a_t)$ incorporates the rewards of the \emph{logged} actions $a_{t+1{:}t+n-1}$, evaluating the initial decision as if the agent would keep executing the behavior policy rather than switch to the improving one. On heterogeneous offline data this induces a systematic pessimistic bias that grows with $n$ and can contribute to the degradation of uncorrected $n$-step TD at long horizons \citep{abraham2026long}. Likelihood-ratio corrections, such as importance sampling (IS),
Retrace, and V-trace
\citep{precup2000eligibility,munos2016safe,espeholt2018impala},
are often difficult to apply to expressive generative policies based
on diffusion or flow matching because exact or inexpensive action
likelihoods may be unavailable
\citep{chi2025diffusion,wang2023diffusion,park2025flow,black2024pi0}. A recent constraint-based alternative, Long-Horizon Q-learning (LQL)
\citep{abraham2026long}, augments a $1$-step critic with
trajectory-level optimality inequalities. Although effective, its
update remains trajectory-structured: the constraints use critic
values at multiple positions within each sampled sequence.

We ask: \emph{can the rapid propagation of $n$-step backups be kept,
and their off-policy bias reduced, using only the two critic
evaluations performed by a standard $n$-step update?} Our answer is
\textbf{Expectile $n$-step Q-learning (ENQ)}. ENQ replaces the
symmetric squared $n$-step TD loss with an asymmetric expectile loss
\citep{newey1987asymmetric} on the action-value TD error. Unlike LQL's trajectory-level constraints, ENQ evaluates the critic
only at the initial logged pair and the final bootstrap state,
requiring neither intermediate critic evaluations nor pairwise
trajectory constraints.

With expectile level $\tau>\tfrac12$, positive TD residuals receive
weight $\tau$ and negative residuals receive weight $1-\tau$, so the
critic emphasizes high-return logged continuations of $(s_t,a_t)$.
Unlike expectile regression on a state-value function
\citep{kostrikov2021offline}, the asymmetric regression is applied
directly to the action-value error of the initial action.
Figure~\ref{fig:intro_figure} summarizes the contrast with one-step
TD, LQL, and uncorrected $n$-step TD.

Our contributions are:
\begin{itemize}
\item \textbf{Simple and flexible update.}
ENQ is a drop-in replacement for symmetric $n$-step TD that requires
no action likelihoods, importance ratios, auxiliary value networks, or
trajectory-level pairwise penalties. It applies to both single-action
and action-chunked critics, is agnostic to the policy-extraction
mechanism, and adds only the expectile level $\tau$.
\item \textbf{Theory.}
(i)~The ENQ operator is a $\gamma^{n}$-contraction for every $\tau$;
(ii)~under deterministic dynamics, the pessimism of $n$-step TD admits
an exact decomposition, the fixed point is nondecreasing in $\tau$ and
bounded above by $Q^*$, and
$(\mathcal T^{n}_{1}Q^*)(s,a)=Q^*(s,a)$ on covered in-support pairs;
(iii)~under the same deterministic-dynamics assumption, the $\tau=1$
fixed point satisfies LQL's lower-bound inequality at separations
$n,2n,\ldots$; and
(iv)~under stochastic dynamics, we derive two-sided operator-bias
bounds and a corresponding fixed-point error bound with
horizon-independent noise constants.
\item \textbf{Cross-task evaluation.}
Using a single expectile level $\tau=0.8$ and a fixed backup horizon $n=4$
across 27 task instances, without task-specific tuning, ENQ is
competitive with LQL on aggregate across OGBench and RoboMimic and
achieves higher measured training throughput.
\end{itemize}
\section{Related Work}

\textbf{Multi-step off-policy learning.} Multi-step backups accelerate reward propagation and remain a primary lever for scaling value learning to long horizons \citep{sutton1988learning,de2018multi,hessel2018rainbow,fedus2020revisiting,park2025horizon}, but inherit the behavior policy's suboptimality off-policy. Importance-weighted corrections such as per-decision IS, Retrace, and V-trace \citep{precup2000eligibility,munos2016safe,espeholt2018impala} require likelihood ratios and can incur high variance, while uncorrected traces require sufficient agreement between the behavior and target policies \citep{kozuno2021revisiting}. Both conditions can be difficult to satisfy with expressive generative actors \citep{chi2025diffusion,park2025flow,li2026reinforcement}. ENQ requires no action likelihoods: the asymmetric loss acts on residuals of sampled $n$-step targets.

\textbf{Optimality inequalities.} Optimality tightening \citep{he2016learning} and LQL \citep{abraham2026long} turn trajectory-wise optimality inequalities into hinge penalties. LQL reuses the online and target critic outputs computed along each sampled trajectory to construct its penalties. ENQ uses a different mechanism: under deterministic dynamics,
Corollary~\ref{cor:implicit} formalizes the connection between the
$\tau=1$ ENQ fixed point and LQL's lower-bound constraints.

\textbf{Critic objectives and sequence models.}
Alternative critic modifications redesign either the learning
objective or the architecture: \citet{farebrother2024stopregressingtrainingvalue}
train value functions through categorical cross-entropy, while
\citet{tian2026chunkingcritictransformerbasedsoft} condition a
Transformer critic on short trajectory segments and aggregated
$n$-step returns. ENQ instead retains a standard state--action critic
and changes the asymmetric weighting of its scalar $n$-step
regression.

\textbf{Asymmetric value learning.}
Implicit Q-Learning (IQL) \citep{kostrikov2021offline} fits a state-value function to the
$\tau$-expectile of $1$-step targets, approximating an in-support
maximum without out-of-distribution queries; Extreme Q-Learning (XQL)
\citep{garg2023extreme} and implicit value regularization
\citep{xu2023offline,xiao2023insample} pursue related in-sample
objectives through other asymmetries. Closely related to ENQ in loss placement, Lower Expectile
Q-learning (LEQ) \citep{park2025modelbased} applies lower-expectile
regression to multi-step $\lambda$-return action-value targets
generated by learned-model rollouts. LEQ uses a lower expectile to
counter model-induced overestimation, whereas ENQ fits an upper
expectile of logged $n$-step targets to counter pessimism caused by
suboptimal behavioral continuations. Asymmetric statistics also underpin distributional RL
\citep{dabney2018distributional,rowland2019statistics}.
Remark~\ref{rem:nstepiql} relates ENQ to an $n$-step analogue of IQL,
which we compare to empirically.

\textbf{Conservatism and pessimism.} A complementary line \emph{induces} pessimism deliberately, to suppress out-of-distribution overestimation: policy constraints and critic penalties \citep{fujimoto2019off,kumar2019stabilizing,kumar2020conservative,fujimoto2021minimalist}, ensemble lower-confidence bounds \citep{an2021uncertainty,bai2022pessimistic,ghasemipour2022so}, and anti-exploration bonuses \citep{rezaeifar2022offline,nikulin2023antiexploration}, with matching statistical guarantees \citep{jin2021pessimism,rashidinejad2021bridging}. That pessimism targets \emph{unsupported} actions; the pessimism ENQ targets is the incidental behavioral bias that multi-step targets impose on \emph{in-support} decisions. The two sources of pessimism are conceptually distinct, although their effects can interact; ENQ retains a conservative ensemble aggregation \eqref{eq:agg} for the former.

\section{Preliminaries}
\label{sec:prelims}

We consider a Markov decision process (MDP) $\mathcal{M}=(\mathcal{S},\mathcal{A},P,R,\gamma)$ with $\gamma\in[0,1)$ and bounded rewards $|R|\le R_{\max}$; write $Q_{\max}\triangleq R_{\max}/(1-\gamma)$. A policy $\pi(\cdot\mid s)$ induces trajectories with $r_t=R(s_t,a_t)$, $s_{t+1}\sim P(\cdot\mid s_t,a_t)$. For every $(s,a)\in\mathcal{S}\times\mathcal{A}$, the optimal action-value
function is the unique bounded solution of
\begin{equation}
\label{eq:bellman}
Q^*(s,a)=R(s,a)+\gamma\,\E_{s'\sim P(\cdot\mid s,a)}\!\left[\max_{a'\in\mathcal{A}}Q^*(s',a')\right],
\end{equation}
and $\supnorm{Q^*}\le Q_{\max}$.
Learning proceeds offline-to-online \citep{nakamoto2023calql,10.1609/aaai.v38i15.29633} from a dataset $\mathcal{D}$ generated by an unknown behavior policy $\pi_\beta$. For the theoretical analysis, we call $(s,a)$ \emph{in-support} if it
belongs to the support of the state--action occupancy measure induced
by $\pi_\beta$.
We sample length-$n$ \emph{segments} $(s_{t:t+n},a_{t:t+n-1},r_{t:t+n-1})$ and write $G_{i:j}\triangleq\sum_{u=i}^{j-1}\gamma^{u-i}r_u$ for partial returns. Standard $n$-step TD trains $Q_\theta$ toward
\begin{equation}
\label{eq:nstep_target}
G_{t:t+n}+\gamma^{n}Q_{\bar\theta}\!\big(s_{t+n},a^*(s_{t+n})\big)
\end{equation}
with a target network $Q_{\bar\theta}$ and a bootstrap action $a^*(s')$ produced by a learned actor $\pi_\phi$ (a sample from $\pi_\phi(\cdot\mid s')$ for stochastic actors), coupling the evaluation of $(s_t,a_t)$ to the logged actions $a_{t+1:t+n-1}\sim\pi_\beta$. Throughout, $\max_{a'}$ abbreviates $\sup_{a'}$; no result requires attainment, and the supremum is attained when $\mathcal{A}$ is compact and $Q(s,\cdot)$ is continuous.\\
\textbf{Horizon notation ($n$ vs.\ $L$).} Throughout, $n$ denotes ENQ's bootstrap horizon: the single temporal separation entering the $n$-step target \eqref{eq:nstep_target}, and hence the contraction modulus $\gamma^{n}$ of the operator analyzed later. We reserve $L$ for the length of the trajectories that LQL \citep{abraham2026long} samples from the replay buffer. LQL trains a \emph{$1$-step} TD critic and adds hinge penalties relating value estimates at pairs of positions within each length-$L$ trajectory, so $L$ indexes a whole family of separations $\{2,\dots,L\}$ rather than a single backup length; the two quantities are not interchangeable. Where we compare bias bounds under stochastic dynamics (Remark~\ref{rem:lqlcompare}), we specialize LQL's longest lower-bound signal to separation $n$, which makes the two violation variables coincide.
\paragraph{LQL optimality inequalities.}
For later comparison, LQL starts from two consequences of optimality.
For any $i<j$, the upper-bound (UB) and lower-bound (LB)
inequalities are
\begin{equation}
\label{eq:lql_inequalities}
\begin{aligned}
\text{(UB)}\quad
\max_{a'}Q^*(s_i,a')
&\ge
\E\!\left[
G_{i:j}
+\gamma^{j-i}Q^*(s_j,a_j)
\right],\\
\text{(LB)}\quad
Q^*(s_i,a_i)
&\ge
\E\!\left[
G_{i:j}
+\gamma^{j-i}\max_{a'}Q^*(s_j,a')
\right].
\end{aligned}
\end{equation}
The expectations are over the transition dynamics and any
stochasticity in the action sequence. LQL replaces them with sampled
trajectory quantities and applies squared hinge penalties to positive
violations; its practical UB family additionally includes the
same-state case $i=j$, for which $G_{i:i}=0$
\citep{abraham2026long}.

\begin{definition}[Expectile \citep{newey1987asymmetric}]
\label{def:expectile}
For a square-integrable random variable $X$ and $\tau\in(0,1)$, the $\tau$-expectile is
$e_\tau[X]\triangleq\arg\min_{m}\E[\ell_\tau(X-m)]$ with the asymmetric squared loss $\ell_\tau(u)\triangleq|\tau-\ind(u<0)|\,u^2$.
\end{definition}

\begin{lemma}[Expectile properties]
\label{lem:expectile}
For bounded $X,Y$ on a common probability space and $c\in\mathbb{R}$:
(i)~$e_{1/2}[X]=\E[X]$;
(ii)~$X\le Y$ a.s.\ $\Rightarrow e_\tau[X]\le e_\tau[Y]$;
(iii)~$e_\tau[X+c]=e_\tau[X]+c$;
(iv)~$\tau\mapsto e_\tau[X]$ is nondecreasing with $e_\tau[X]\to\esssup X$ as $\tau\to1$;
(v)~$|X-Y|\le c$ a.s.\ $\Rightarrow|e_\tau[X]-e_\tau[Y]|\le c$.
Properties (ii)--(v) extend to $e_1[X]\triangleq\esssup X$. (Proof: Appendix~\ref{app:expectile}.)
\end{lemma}

\section{Expectile $n$-step Q-learning}
\label{sec:method}

The core modification relative to $n$-step TD is the replacement of the symmetric squared loss on the TD error by the asymmetric loss $\ell_\tau$ of Definition~\ref{def:expectile}; no additional value or correction networks, trajectory penalties, or action likelihoods are introduced. We describe the single-action critic; the action-chunked variant replaces actions by chunks of $h$ consecutive actions \citep{li2026reinforcement} and is otherwise identical.

\paragraph{Target.} Given a segment from $\mathcal{D}$ and a critic ensemble $\{Q_{\theta,i}\}_{i=1}^{K}$, the bootstrap aggregates the target ensemble with an optional conservative penalty $\rho\ge0$ \citep{ghasemipour2022so},
\begin{equation}
\label{eq:agg}
\bar Q^{\rho}_{\bar\theta}(s,a)=\tfrac{1}{K}\textstyle\sum_{i}Q_{\bar\theta,i}(s,a)-\rho\operatorname{std}_{i}Q_{\bar\theta,i}(s,a),
\end{equation}
yielding the per-segment target
\begin{equation}
\label{eq:enq_target}
Y_{t,n}=G_{t:t+n}+\gamma^{n}\,m_{t+n}\,\bar Q^{\rho}_{\bar\theta}\!\big(s_{t+n},a^*(s_{t+n})\big),
\end{equation}
where $m_{t+n}\in\{0,1\}$ masks the bootstrap at terminal states and a validity mask $v_t=\prod_{k=t}^{t+n-2}(1-d_k)$ ($d_k$ indicating termination at step $k$) discards segments crossing episode boundaries.
\paragraph{Loss.} With per-critic errors $\delta_i\triangleq \sg[Y_{t,n}]-Q_{\theta,i}(s_t,a_t)$ ($\sg$: stop-gradient),
\begin{equation}
\label{eq:enq_loss}
\mathcal{L}_{\mathrm{ENQ}}(\theta)=\E_{\mathcal{D}}\!\Big[\,v_t\cdot\tfrac{1}{K}\textstyle\sum_{i=1}^{K}\ell_\tau(\delta_i)\Big].
\end{equation}
At $\tau=\tfrac12$, the expectile loss is proportional to the
squared-error $n$-step TD loss and has the same population minimizer; for $\tau>\tfrac12$, positive residuals receive weight $\tau$ and negative residuals weight $1-\tau$, so the critic tracks the $\tau$-expectile of the $n$-step target distribution conditioned on $(s_t,a_t)$, the remaining randomness being the logged actions $a_{t+1:t+n-1}\sim\pi_\beta$ and, in stochastic environments, the dynamics. At the population level this continuation randomness is aggregated through the expectile and the resulting statistic is assigned to $Q(s_t,a_t)$; each sampled target still depends on the realized logged continuation. Pseudocode is given in Appendix~\ref{app:algorithm} (Algorithm~\ref{alg:enq}).
\paragraph{Policy extraction.}
ENQ is agnostic to how a policy is extracted from $Q_\theta$; we
instantiate Best-of-$N$ reranking, Flow Q-Learning (FQL) distillation,
and chunked FQL, following the policy-extraction procedures of
\citet{park2025flow,li2026reinforcement,abraham2026long}. This modularity is useful because policy extraction can itself be a
major determinant of offline RL performance and scalability
\citep{park2024value}.

\paragraph{Computational profile.}
ENQ evaluates the online critic once at the initial pair $(s_t,a_t)$
and the target critic once at the bootstrap endpoint, then applies a
scalar expectile regression to the resulting $n$-step residual. LQL
instead evaluates online and target critics at every position of a
length-$L$ trajectory and constructs lower- and upper-bound hinge
penalties over multiple pairs of positions
\citep{abraham2026long}. Although LQL reuses these trajectory-wise
outputs across its TD and hinge terms, its update still requires more
critic evaluations and pairwise constraint computations than ENQ.
We quantify the resulting difference through training throughput under
matched horizon, batch size, hardware, and software settings.
\section{Theoretical Analysis}
\label{sec:theory}

We analyze the exact operator underlying \eqref{eq:enq_loss}. Three idealizations apply: the population expectile replaces the sampled loss, the greedy bootstrap replaces the actor proxy, and function approximation is not modeled; Appendix~\ref{app:scope} discusses their implications, and similar idealizations appear in prior analyses of IQL and LQL. Throughout, $\supnorm{\cdot}$ is the sup norm over $\mathcal{S}\times\mathcal{A}$ and all value functions satisfy $\supnorm{Q}\le Q_{\max}$.

For bounded $Q$, define the random $n$-step target seeded at $(s,a)$,
\begin{equation}
\label{eq:Z}
\begin{split}
\Zn{Q}(s,a)\triangleq\;&\textstyle\sum_{k=0}^{n-1}\gamma^{k}r(s_k,a_k)+\gamma^{n}\max_{a'}Q(s_n,a'),\\[-2pt]
&\ s_0{=}s,\ a_0{=}a,\ a_k\sim\pi_\beta(\cdot\mid s_k),\ 1{\le}k{\le}n{-}1,
\end{split}
\end{equation}
with $s_{k+1}\sim P(\cdot\mid s_k,a_k)$. Let $\Prob_{s,a}$ denote the law of the segment $(s_0,a_0,\dots,s_n)$ so generated; all expectations, variances, expectiles, and essential suprema over segment quantities below are taken under $\Prob_{s,a}$, written $\E_{s,a}$, $\mathrm{Var}_{s,a}$ where the conditioning is not clear from context. The \emph{expectile $n$-step Bellman operator} is $(\Top Q)(s,a)\triangleq e_\tau[\Zn{Q}(s,a)]$.

\begin{theorem}[Contraction]
\label{thm:contraction}
For every $\tau\in(0,1]$ and $n\ge1$, and arbitrary (stochastic) dynamics and behavior policies,
$\supnorm{\Top Q_1-\Top Q_2}\le\gamma^{n}\supnorm{Q_1-Q_2}$.
Hence $\Top$ has a unique bounded fixed point $\Qfix$, reached by iteration at geometric rate $\gamma^{n}$ per application.
\end{theorem}

Theorem~\ref{thm:contraction} makes ENQ well-posed for every $\tau$, with contraction modulus $\gamma^{n}$ versus $\gamma$ for $1$-step methods. To locate the fixed point, define the \emph{suboptimality gap} $\Delta^*(s,a)\triangleq\max_{a'}Q^*(s,a')-Q^*(s,a)\ge0$ and the \emph{one-step Bellman noise} $\epsilon_k\triangleq r(s_k,a_k)+\gamma\max_{a'}Q^*(s_{k+1},a')-Q^*(s_k,a_k)$, which satisfies $\E[\epsilon_k\mid s_k,a_k]=0$ by \eqref{eq:bellman}. Pathwise (Appendix~\ref{app:telescope}),
\begin{align}
W_n(s,a) &\triangleq \Zn{Q^*}(s,a)-Q^*(s,a)=\sum_{k=0}^{n-1}\gamma^{k}\epsilon_k \notag\\
&\quad{}-\sum_{k=1}^{n-1}\gamma^{k}\Delta^*(s_k,a_k), \label{eq:telescope}
\end{align}
a mean-zero noise term minus a nonnegative behavioral-drift term with mean $\mu_n(s,a)\triangleq\sum_{k=1}^{n-1}\gamma^{k}\E_{s,a}[\Delta^*(s_k,a_k)]$.

\begin{assumption}[Deterministic dynamics]
\label{ass:det}
$s_{k+1}=f(s_k,a_k)$; then $\epsilon_k\equiv0$ and all randomness in $\Zn{Q}$ comes from $\pi_\beta$.
\end{assumption}

\begin{assumption}[Optimal coverage]
\label{ass:coverage}
For every in-support $(s,a)$,
\[
\operatorname*{ess\,inf}_{\Prob_{s,a}}
\sum_{k=1}^{n-1}\gamma^{k}\Delta^*(s_k,a_k)=0.
\]
\end{assumption}
Assumption \ref{ass:coverage} implies that the behavior policy places positive probability on continuations whose cumulative optimality gap is arbitrarily small; a segment-level analogue of the optimal-policy coverage conditions of offline-RL theory \citep{rashidinejad2021bridging,jin2021pessimism}.
\begin{theorem}[Pessimism decomposition and endpoint correction]
\label{thm:bias}
Under Assumption~\ref{ass:det}:
\textbf{(a)}~$(\mathcal T^{n}_{1/2}Q^*)(s,a)-Q^*(s,a)=-\mu_n(s,a)\le0$, and $\mu_n$ is nondecreasing in $n$;
\textbf{(b)}~$\Qfix$ is pointwise nondecreasing in $\tau$;
\textbf{(c)}~under Assumption~\ref{ass:coverage},
$(\mathcal T^{n}_{1}Q^*)(s,a)=Q^*(s,a)$ for every in-support $(s,a)$;

\textbf{(d)}~for every $\tau\in(0,1]$,
$(\Top Q^*)(s,a)\le Q^*(s,a)$, and hence
$\Qfix\le Q^*$ pointwise.

\end{theorem}
Since $e_{1/2}$ is the conditional mean, $\mathcal T^{n}_{1/2}$ is the standard $n$-step TD operator, and (a) quantifies its pessimism as one discounted suboptimality gap per intervening logged action. By (b) and (d), $\Qfix$ is nondecreasing in $\tau$ and never exceeds $Q^*$. Part~(c) shows that the one-application operator bias at $Q^*$ vanishes at the $\tau=1$ endpoint on covered in-support pairs; it does not identify the global fixed point with $Q^*$. This analysis motivates examining how the empirically preferred $\tau$ varies with the horizon and data quality; ablation~(A1) probes this question.

\begin{corollary}[Implicit lower-bound satisfaction]
\label{cor:implicit}
Under Assumption~\ref{ass:det}, for every $j\ge1$ and $\Prob_{s_t,a_t}$-almost every logged length-$jn$ segment seeded at an in-support $(s_t,a_t)$,
\[
Q^{n}_{1}(s_t,a_t)\;\ge\;G_{t:t+jn}+\gamma^{jn}\max_{a'}Q^{n}_{1}(s_{t+jn},a').
\]
\end{corollary}

The display is the separation-$jn$ instance of the lower-bound
inequality (LB) in \eqref{eq:lql_inequalities}. LQL penalizes sampled
violations across multiple separations within each length-$L$
trajectory \citep{abraham2026long}, whereas the $\tau=1$ endpoint
fixed point satisfies the separation-$jn$ instances by construction.

The upper-bound inequality (UB) in
\eqref{eq:lql_inequalities} has no ENQ analogue; the conservative aggregation \eqref{eq:agg} is a counterweight of a different kind, and Theorem~\ref{thm:stoch} indicates when it may help: for large $\tau$, the expectile places increasing weight on favorable transition realizations, shifting the target upward relative to its conditional mean and potentially inducing positive operator bias.
Let $\sigma_n^2(s,a)\triangleq\mathrm{Var}_{s,a}(W_n(s,a))$; by the moment bounds of Appendix~\ref{app:bias}, $\sigma_n^2\le\Vinf\triangleq M^2/(1-\gamma)^2$ with $M\triangleq R_{\max}+(1+\gamma)Q_{\max}+2Q_{\max}$, the constants of LQL's stochasticity analysis. Write $\kt\triangleq\frac{2\tau-1}{2(1-\tau)}$ and $B_\tau(s,a)\triangleq(\Top Q^*)(s,a)-Q^*(s,a)=e_\tau[W_n(s,a)]$.

\begin{theorem}[Bias under stochastic dynamics]
\label{thm:stoch}
For any dynamics and $\tau\in[\tfrac12,1)$:
\textbf{(a)}~$-\mu_n(s,a)\;\le\;B_\tau(s,a)\;\le\;-\mu_n(s,a)+\kt\,\sigma_n(s,a)$;
\textbf{(b)}~If $\mu_n(s,a)+\sigma_n(s,a)>0$, then $B_\tau(s,a)\le0$ whenever
\[
\tau\le\tau_{\mathrm{safe}}(s,a)
\triangleq
\frac{\sigma_n(s,a)+2\mu_n(s,a)}{2\sigma_n(s,a)+2\mu_n(s,a)}.
\]
When $\mu_n(s,a)=\sigma_n(s,a)=0$, we set $\tau_{\mathrm{safe}}(s,a)=1$; in this case $B_\tau(s,a)=0$ for every $\tau$.
\textbf{(c)}~$\supnorm{\Qfix-Q^*}\le\big(\sup_{s,a}\mu_n(s,a)+\kt\sqrt{\Vinf}\big)/(1-\gamma^{n})$.
\end{theorem}

In (a), the pessimistic side is the drift of Theorem~\ref{thm:bias}(a), while the optimistic side is at most $\kt\sqrt{\Vinf}$, independent of $n$. For $\sigma_n>0$, the threshold in (b) increases with the
drift-to-variability ratio $\mu_n/\sigma_n$. For fixed $\mu_n>0$, it
tends to $1$ as $\sigma_n\to0$. This sufficient condition is generally
conservative and does not subsume Theorem~\ref{thm:bias}(d), which uses
the stronger pathwise sign of $W_n$ under deterministic dynamics. Since $\mu_n$ and $\sigma_n$ are unobservable, the bound suggests only that conservative aggregation may help when upper-tail emphasis is large relative to the behavioral drift; ablation~(A4) probes this interaction in one controlled setting.

\begin{remark}[Comparison with LQL's stochasticity bound]
\label{rem:lqlcompare}
$W_n$ coincides with the lower-bound violation signal of \citet{abraham2026long} (LQL's Appendix~C.1, $L$ set to $n$), analyzed with the same constants $M$ and $\Vinf$. LQL bounds spurious penalty activations; Theorem~\ref{thm:stoch} bounds the operator bias and, because ENQ is a single contraction, converts it into a fixed-point error bound. Moreover, $\tau$ enters ENQ's bound, whereas LQL's hinge weights do not enter theirs.
\end{remark}

Finally, why place the expectile on the action-value error rather than on a state-value function as in IQL? A natural $n$-step analogue of IQL is
\begin{equation}
\label{eq:nstepiql}
Q(s_t,a_t)\leftarrow G_{t:t+n}+\gamma^{n}V(s_{t+n}),
\end{equation}
where $V$ is fit by expectile regression on $n$-step returns.

\begin{remark}[$n$-step IQL-style baseline]
\label{rem:nstepiql}
Equation~\eqref{eq:nstepiql} gives a practical baseline that routes
the asymmetry through an auxiliary state-value function rather than
applying it directly to the action-value residual. This introduces an
additional coupled value regression; related work uses a
$V$-ensemble to suppress accumulation of initial value errors in
IQL-style learning \citep{chen2025active}. We compare the baseline
with ENQ under transferred ENQ hyperparameters in ablation~(A3).
\end{remark}

\section{Experiments}
\label{sec:experiments}

\begin{table*}[t]
  \vspace{-1\baselineskip}
\caption{\textbf{OGBench.} Final success rate (\%) on four domains,
each containing five tasks, with four seeds per task. Values are
unweighted task means; brackets give $95\%$ bootstrap confidence
intervals for ENQ. Baseline values are from
\citet{li2026reinforcement,abraham2026long}; Revisited
Behavior-Regularized Actor-Critic (ReBRAC) was introduced by
\citet{tarasov2023rebrac}. Throughout the result tables, bold marks
values at least $0.95$ times the row maximum; overbars mark values at
least $0.95$ times the maximum within the corresponding FQL or
action-chunking block.}
  \centering
  \begin{small}
  \resizebox{\textwidth}{!}{%
  \setlength{\tabcolsep}{4pt}
  \renewcommand{\arraystretch}{1.7}
    \begin{tabular}{l|cccc|ccc|cc}
      \toprule
       & \multicolumn{4}{c|}{\textbf{FQL}} & \multicolumn{3}{c|}{\textbf{Action Chunking}} & \multicolumn{2}{c}{\textbf{Other baselines}} \\
        \cmidrule(lr){2-5} \cmidrule(lr){6-8} \cmidrule(lr){9-10}
       & TD & TD-$n$ & LQL & ENQ & TD & LQL & ENQ & ReBRAC & IQL \\
      \midrule
      \texttt{cube-double} & 18.3 & 49.8 & \barnum{86.2} & \makecell[tc]{\barnum{84.5} \\[-3pt] {\scriptsize \textcolor{black!55}{[78.8, 93.0]}}} & \barnum{\textbf{94.2}} & \barnum{\textbf{95.1}} & \makecell[tc]{\barnum{\textbf{97.8}} \\[-3pt] {\scriptsize \textcolor{black!55}{[97.3, 98.4]}}} & 30 & 0 \\
      \texttt{cube-triple} & 9.3 & 6.9 & 31.2 & \makecell[tc]{\barnum{40.3} \\[-3pt] {\scriptsize \textcolor{black!55}{[37.6, 43.1]}}} & 35.2 & \barnum{\textbf{52.1}} & \makecell[tc]{\barnum{\textbf{51.2}} \\[-3pt] {\scriptsize \textcolor{black!55}{[48.9, 54.2]}}} & 0 & 0 \\
      \texttt{humanoid-md} & 40.4 & 28.6 & \barnum{\textbf{65.2}} & \makecell[tc]{59.8 \\[-3pt] {\scriptsize \textcolor{black!55}{[59.5, 59.9]}}} & 15.4 & 22.2 & \makecell[tc]{\barnum{34.1} \\[-3pt] {\scriptsize \textcolor{black!55}{[29.6, 37.0]}}} & 3.5 & 23.1 \\
      \texttt{antmaze-giant} & 4.8 & 28.8 & 53.3 & \makecell[tc]{\barnum{\textbf{91.4}} \\[-3pt] {\scriptsize \textcolor{black!55}{[79.2, 96.8]}}} & 22.2 & \barnum{40.7} & \makecell[tc]{\barnum{40.2} \\[-3pt] {\scriptsize \textcolor{black!55}{[29.5, 53.6]}}} & 57.1 & 3.2 \\
      \midrule
      \textbf{Total} & 18.2 & 28.5 & 59.0 & \makecell[tc]{\barnum{\textbf{69.0}} \\[-3pt] {\scriptsize \textcolor{black!55}{[65.9, 71.5]}}} & 41.8 & 52.5 & \makecell[tc]{\barnum{55.9} \\[-3pt] {\scriptsize \textcolor{black!55}{[52.9, 59.4]}}} & 22.7 & 6.6 \\
      \bottomrule
    \end{tabular}%
  }
    \label{table:OGBench_success}
  \end{small}
\end{table*}

\begin{table}[ht]
\caption{\textbf{\texttt{humanoidmaze-giant}.} Final success rate
(\%) on five tasks with four seeds per task. Brackets give $95\%$
bootstrap confidence intervals for the aggregate row. Baseline results
are from \citet{abraham2026long}; policy extraction uses Best-of-$N$.}
  \centering
  \begin{small}
  \setlength{\tabcolsep}{4pt}
  \renewcommand{\arraystretch}{1.7}
    \begin{tabular}{lcccc}
      \toprule
       & TD & TD-$4$ & LQL ($L=64$) & ENQ (ours) \\
      \midrule
      \texttt{task1} & 0.0 & 18.0 & 31.3 & \textbf{95.0} \\
      \texttt{task2} & 0.0 & 63.3 & \textbf{97.3} & \textbf{97.7} \\
      \texttt{task3} & 0.0 & 7.3 & 70.7 & \textbf{79.7} \\
      \texttt{task4} & 0.0 & 4.7 & \textbf{80.7} & 32.3 \\
      \texttt{task5} & 0.0 & \textbf{98.7} & \textbf{98.7} & \textbf{99.7} \\
      \midrule
      \textbf{Total} & \makecell[tc]{0.0 \\[-3pt] {\scriptsize \textcolor{black!55}{[0,0]}}} & \makecell[tc]{38.4 \\[-3pt] {\scriptsize \textcolor{black!55}{[35,41]}}} & \makecell[tc]{75.7 \\[-3pt] {\scriptsize \textcolor{black!55}{[70,81]}}} & \makecell[tc]{\textbf{80.9} \\[-3pt] {\scriptsize \textcolor{black!55}{[74,94]}}} \\
      \bottomrule
    \end{tabular}
    \label{table:hmgiant_success}
  \end{small}
\end{table}

\begin{table}[t]
\caption{\textbf{RoboMimic.} Final success rate (\%) on two tasks
with four seeds per task. Brackets give $95\%$ bootstrap confidence
intervals. Baseline results are from \citet{abraham2026long}.}
  \centering
  \begin{small}
  \setlength{\tabcolsep}{5pt}
  \renewcommand{\arraystretch}{1.7}
    \begin{tabular}{l|ccc|cc}
      \toprule
       & \multicolumn{3}{c|}{\textbf{FQL}} & \multicolumn{2}{c}{\textbf{Action Chunking}} \\
        \cmidrule(lr){2-4} \cmidrule(lr){5-6}
       & TD-$n$ & LQL & ENQ & LQL & ENQ \\
      \midrule
      \texttt{square} & \makecell[tc]{15.5 \\[-3pt] {\scriptsize \textcolor{black!55}{[8,22]}}} & \makecell[tc]{\barnum{\textbf{61.0}} \\[-3pt] {\scriptsize \textcolor{black!55}{[60,63]}}} & \makecell[tc]{\barnum{\textbf{59.0}} \\[-3pt] {\scriptsize \textcolor{black!55}{[54,64]}}} & \makecell[tc]{23.5 \\[-3pt] {\scriptsize \textcolor{black!55}{[16,31]}}} & \makecell[tc]{\barnum{41.3} \\[-3pt] {\scriptsize \textcolor{black!55}{[38,43]}}} \\
      \texttt{can} & \makecell[tc]{69.5 \\[-3pt] {\scriptsize \textcolor{black!55}{[63,73]}}} & \makecell[tc]{\barnum{\textbf{90.0}} \\[-3pt] {\scriptsize \textcolor{black!55}{[84,96]}}} & \makecell[tc]{\barnum{\textbf{89.0}} \\[-3pt] {\scriptsize \textcolor{black!55}{[84,93]}}} & \makecell[tc]{\barnum{\textbf{85.5}} \\[-3pt] {\scriptsize \textcolor{black!55}{[80,90]}}} & \makecell[tc]{\barnum{83.5} \\[-3pt] {\scriptsize \textcolor{black!55}{[80,88]}}} \\
      \midrule
      \textbf{Total} & \makecell[tc]{42.5 \\[-3pt] {\scriptsize \textcolor{black!55}{[38,47]}}} & \makecell[tc]{\barnum{\textbf{75.5}} \\[-3pt] {\scriptsize \textcolor{black!55}{[72,78]}}} & \makecell[tc]{\barnum{\textbf{74.0}} \\[-3pt] {\scriptsize \textcolor{black!55}{[70,77]}}} & \makecell[tc]{54.5 \\[-3pt] {\scriptsize \textcolor{black!55}{[50,59]}}} & \makecell[tc]{\barnum{62.4} \\[-3pt] {\scriptsize \textcolor{black!55}{[60,65]}}} \\
      \bottomrule
    \end{tabular}
    \label{table:robomimic_success}
  \end{small}
\end{table}

The experiments address four questions motivated by the theoretical and computational analysis: \textbf{(Q1)}~how ENQ compares to $1$-step TD, $n$-step TD, and LQL across long-horizon tasks and policy classes; \textbf{(Q2)}~whether $\tau$ acts as the bias-control parameter of Theorem~\ref{thm:bias}, including its interplay with the horizon $n$; \textbf{(Q3)}~how ENQ and LQL scale with the critic ensemble, and how the conservative coefficient $\rho$ interacts with the expectile level (Theorem~\ref{thm:stoch}); \textbf{(Q4)}~whether placing the asymmetry on the action-value error rather than routing it through an auxiliary state-value function matters in practice (Remark~\ref{rem:nstepiql}).

\textbf{Setup.}
We evaluate 25 task instances from OGBench
\citep{park2025ogbench} and two from RoboMimic
\citep{robomimic2021}. Unless stated otherwise, all ENQ results use
four seeds per task, a fixed expectile level $\tau=0.8$, and a single
fixed backup horizon $n=4$ across all 27 tasks. Across ENQ--LQL performance comparisons, we match the number of
sampled transitions processed per update by choosing batch sizes such
that
\[
B_{\mathrm{ENQ}}n=B_{\mathrm{LQL}}L.
\]
In the default configuration, LQL uses trajectory length $L=8$ and
batch size $128$, while ENQ uses backup horizon $n=4$ and batch size
$256$, so both methods process $1024$ transitions per update. On
\texttt{humanoidmaze-giant}, LQL uses $L=64$ and batch size $128$,
while ENQ uses $n=4$ and batch size $2048$, so both methods process
$8192$ transitions per update. The profiling experiment instead uses
identical horizon length and batch size for both methods to isolate
differences in critic-update computation.

Unless otherwise
stated in the corresponding caption, uncertainty intervals are $95\%$
bootstrap confidence intervals. Further experimental details are
deferred to Appendix~\ref{app:expdetails}.

\subsection{Comparative Evaluation}
\label{subsec:main_results}

Across Tables~\ref{table:OGBench_success},
\ref{table:robomimic_success}, and~\ref{table:hmgiant_success}, ENQ
attains the higher aggregate mean in four of five comparisons, with
LQL ahead only on RoboMimic FQL. The strongest ENQ gains occur on
\texttt{cube-triple}, \texttt{humanoid-md}, and
\texttt{antmaze-giant}, aligning with the trajectory-stitching
structure of these domains \citep{park2025ogbench,ghugare2024closing}. At finer
granularity, LQL remains stronger on \texttt{humanoid-md} with FQL
and on \texttt{humanoidmaze-giant} task~4.

\subsection{Ablations}
\label{subsec:ablations}

\begin{figure}[h]
  \centering
  \includegraphics[width=\columnwidth]{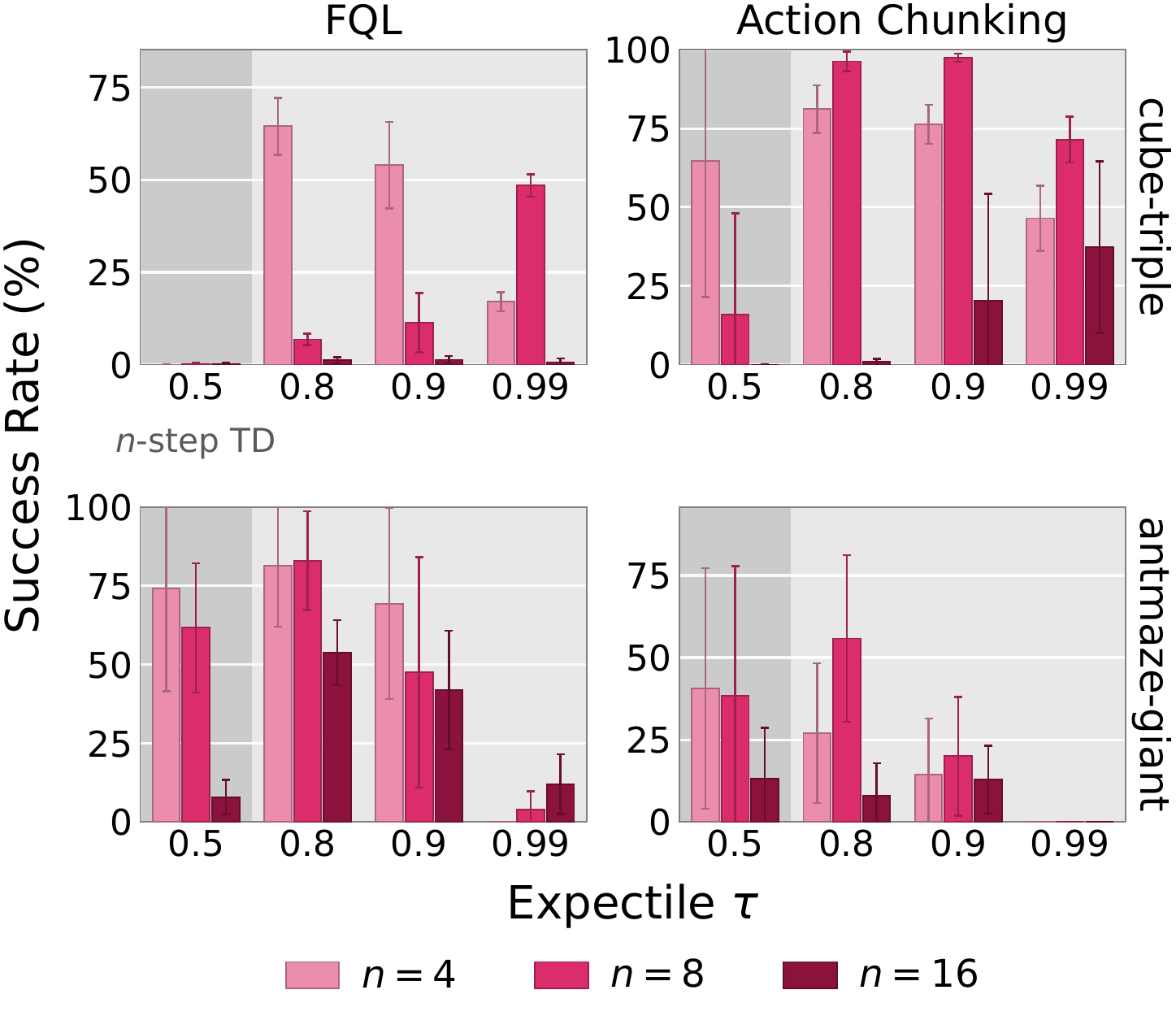}
\caption{Final success across two tasks and two policy-extraction mechanisms for $\tau\in\{0.5,0.8,0.9,0.99\}$ and $n\in\{4,8,16\}$, using four seeds per setting. Error bars show one standard deviation.}
\label{fig:expectile_horizon}
\end{figure}

\textbf{(A1) Expectile level and backup horizon.}
In Figure~\ref{fig:expectile_horizon}, upper expectiles attain the
highest mean in 10 of the 12 task--actor--horizon configurations,
using $\tau=0.5$ as the matched $n$-step TD baseline. The gains can be
large: on \texttt{cube-triple} with FQL, success rises from $0.0\%$
to $64.5\%$ at $n=4$, while on \texttt{antmaze-giant} with FQL it
rises from $7.8\%$ to $53.8\%$ at $n=16$. The effect is nevertheless
non-monotonic: the extreme setting $\tau=0.99$ is brittle on
\texttt{antmaze-giant}, where the symmetric baseline remains
competitive in several action-chunking configurations.

The preferred $\tau$ tends to increase with $n$ on
\texttt{cube-triple}, but no corresponding ordering appears on
\texttt{antmaze-giant}, motivating joint selection of $\tau$ and
$n$. An intermediate horizon performs best overall: three of the four
task--actor combinations peak at $n=8$, whereas $n=16$ is never
optimal. Longer multi-step returns can become variance-limited
\citep{3692070.3692463}, potentially compounding the finite-sample
sensitivity of high expectiles and helping explain the deterioration
at $n=16$.

\begin{figure}[h]
  \centering
  \begin{minipage}[b]{0.48\columnwidth}
    \centering
    \includegraphics[width=\linewidth]{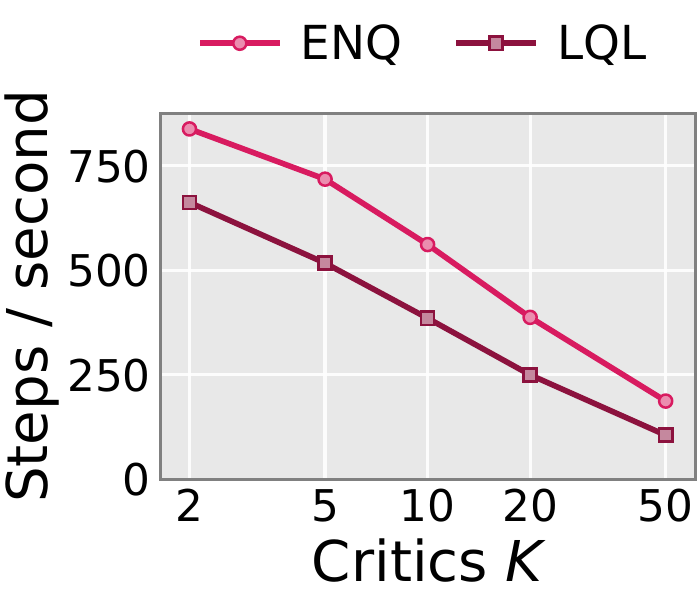}
  \end{minipage}\hfill
  \begin{minipage}[b]{0.48\columnwidth}
    \centering
    \includegraphics[width=\linewidth]{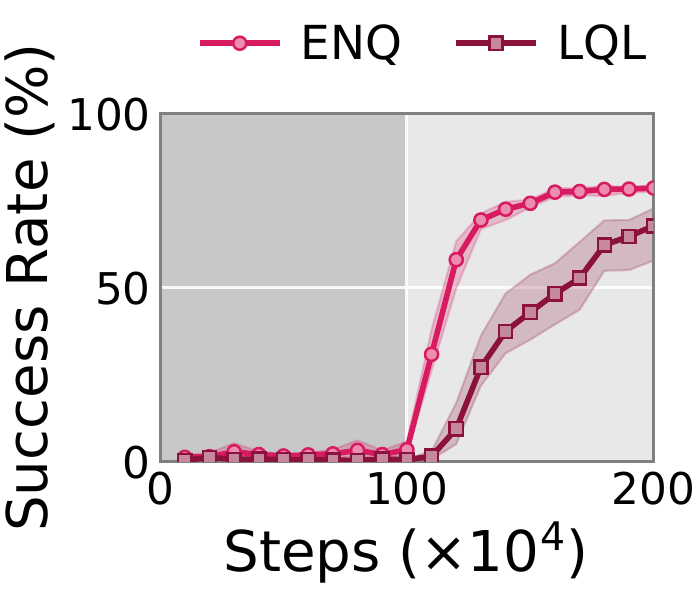}

  \end{minipage}
\caption{\emph{Left:} Training throughput on
\texttt{antmaze-giant} across critic-ensemble sizes, with ENQ and LQL
using identical batch size $B=128$ and horizon length $n=L=8$.
\emph{Right:} Final success on the five \texttt{cube-triple} tasks
with four seeds per task. Shading shows $95\%$ bootstrap confidence
intervals.}
  \label{fig:ensemble_scaling}
\end{figure}

\textbf{(A2) Ensemble scaling: computation and performance.}
Motivated by work treating critic capacity and compute allocation as
scaling axes in value-based RL
\citep{chen2021randomized,nauman2024bigger,rybkin2025valuebased},
Figure~\ref{fig:ensemble_scaling} evaluates ensemble size through
training throughput and final performance. ENQ's throughput advantage
over LQL widens from $1.27\times$ at $K=2$ to $1.77\times$ at
$K=50$.

We compare performance at $K=2$ and $K=10$ while holding all other
hyperparameters fixed. At $K=2$, the corresponding
\texttt{cube-triple} action-chunking results in
Table~\ref{table:OGBench_success} are nearly matched: $51.2\%$ for
ENQ and $52.1\%$ for LQL. At $K=10$, ENQ reaches $78.0\%$, compared
with $59.1\%$ for LQL, with non-overlapping bootstrap confidence
intervals. The separation emerges during online fine-tuning,
indicating that ENQ benefits more from the larger ensemble in this
controlled comparison. We choose $K=10$ to match the large-ensemble
configuration used in recent generative-policy value-learning work
such as Q-Learning with Adjoint Matching
(QAM)~\citep{li2026qlearningadjointmatching}.

\begin{figure}[h]
  \centering
  \begin{minipage}[b]{0.48\columnwidth}
    \centering
    \includegraphics[width=\linewidth]{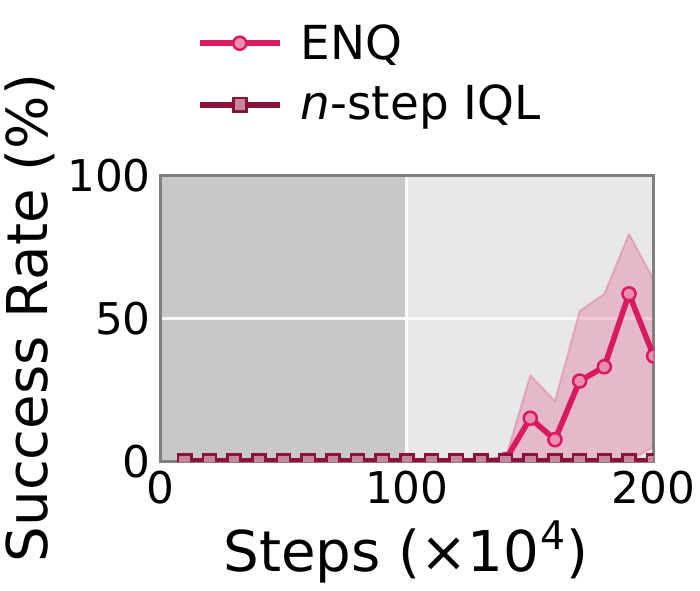}
  \end{minipage}\hfill
  \begin{minipage}[b]{0.48\columnwidth}
    \centering
    \includegraphics[width=\linewidth]{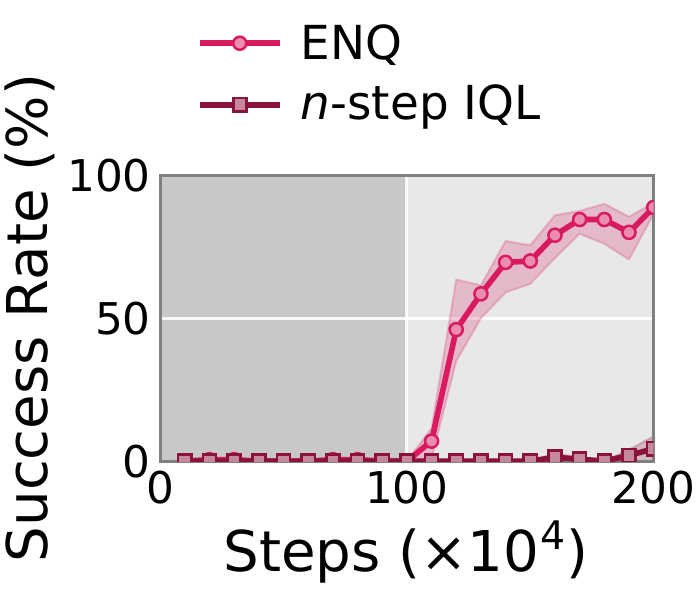}

  \end{minipage}
\caption{Offline-to-online success for ENQ and the $n$-step IQL-style baseline, with action-chunked policy extraction on two tasks, using four seeds per task and transferred ENQ hyperparameters. Shading shows $95\%$ bootstrap confidence intervals.}
\label{fig:expectile_placement}
\end{figure}

\textbf{(A3) Expectile placement.}
Figure~\ref{fig:expectile_placement} shows that ENQ substantially
outperforms the $n$-step IQL-style baseline on both tasks, supporting
the direct placement of expectile weighting on the action-value
residual rather than through an auxiliary state-value regression.

\begin{figure}[h]
  \centering
  \begin{minipage}[b]{0.48\columnwidth}
    \centering
    \includegraphics[width=\linewidth]{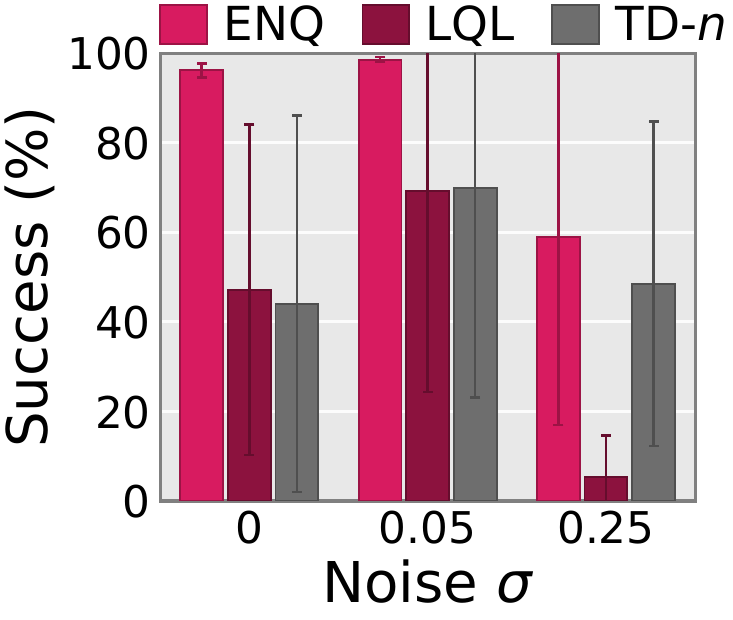}
  \end{minipage}\hfill
  \begin{minipage}[b]{0.48\columnwidth}
    \centering
    \includegraphics[width=\linewidth]{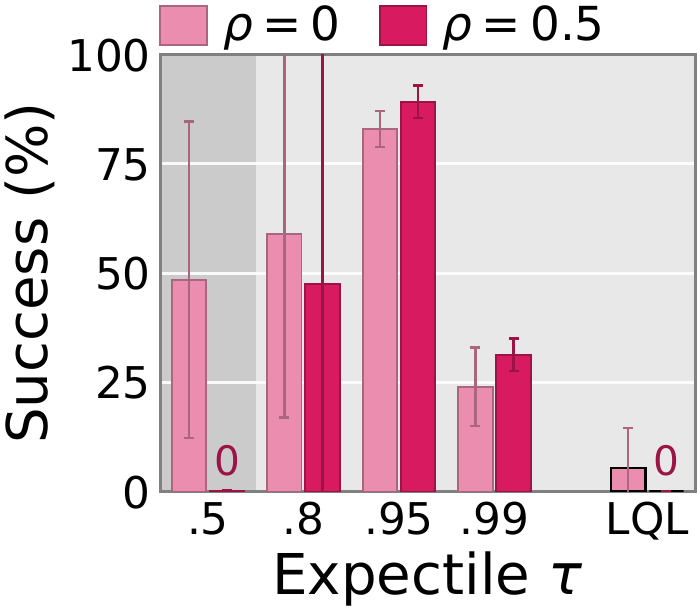}

  \end{minipage}
\caption{Stochastic-transition ablation on
\texttt{cube-triple} task~2 with action-chunked policy extraction,
using four seeds per setting. \emph{Left:} final success of TD-$n$
($\tau=0.5$), LQL, and ENQ ($\tau=0.8$) for
$\sigma\in\{0,0.05,0.25\}$. \emph{Right:} ENQ success at
$\sigma=0.25$ across expectile levels $\tau$ and conservative
coefficients $\rho$. Error bars show $95\%$ bootstrap confidence
intervals.}
\label{fig:transition_noise}
\end{figure}

\textbf{(A4) Robustness to stochastic transitions.}
In Figure~\ref{fig:transition_noise}, ENQ attains the highest mean
success at every tested noise level. Under the strongest noise,
$\sigma=0.25$, it retains $59.0\%$ success, compared with $48.4\%$
for TD-$n$ and $5.4\%$ for LQL. At this noise level, the smaller gain
of the default $\tau=0.8$ setting over TD-$n$ motivates the joint
$\tau$--$\rho$ sweep in the right panel.

At $\sigma=0.25$, conservative aggregation helps primarily at high
expectile levels. Among the tested settings, $\tau=0.95$ with
$\rho=0.5$ performs best at $89.3\%$, whereas adding the same penalty
hurts the lower expectiles. This interaction aligns with the role of
$\rho$ as a counterweight to upper-tail emphasis.

\section{Conclusion}
We introduced ENQ, a simple modification of off-policy $n$-step
Q-learning that applies an upper-expectile loss directly to the
action-value residual. ENQ reduces the influence of poor logged
continuations without action likelihoods, auxiliary value networks, or
trajectory-level pairwise penalties. The resulting operator is a
$\gamma^n$-contraction; under deterministic dynamics, its pessimistic
bias admits an exact decomposition and its $\tau=1$ fixed point
satisfies LQL's lower-bound inequality at separations $n,2n,\ldots$.
Under stochastic dynamics, we derive two-sided operator-bias bounds and
a corresponding fixed-point error bound.

Across 27 manipulation and navigation task instances, ENQ attains the
higher aggregate mean in four of five comparisons with LQL. Its critic
update also achieves $1.27$--$1.77\times$ higher measured training
throughput in the matched profiling study and benefits more strongly
from increasing the critic-ensemble size in the controlled scaling
experiment.

Several questions remain open at the theoretical level. Our analysis
characterizes the population ENQ operator with an exact greedy
bootstrap, and therefore does not model function approximation,
target-network lag, actor approximation, or finite-sample
optimization. Consequently, the contraction and bias guarantees do
not yet quantify the estimation error of practical neural critics.
Extending the fixed-point analysis to approximate operators and
deriving finite-sample guarantees for expectile $n$-step regression
are natural directions for future work.

\bibliography{aaai2027}

\clearpage
\appendix
\setcounter{secnumdepth}{1}

\section*{Technical Appendix}

\section{Expectile properties (Lemma~\ref{lem:expectile})}
\label{app:expectile}

Let $g(m)\triangleq\tau\E[(X-m)_+]-(1-\tau)\E[(m-X)_+]$. Differentiating the objective of Definition~\ref{def:expectile} under the expectation gives $\frac{d}{dm}\E[\ell_\tau(X-m)]=-2g(m)$ (the quadratic hinge is $C^1$). For $m_1<m_2$, the pointwise identity $[(x-m_1)_+-(x-m_2)_+]+[(m_2-x)_+-(m_1-x)_+]=m_2-m_1$, with both brackets in $[0,m_2-m_1]$, yields $g(m_1)-g(m_2)\ge\min(\tau,1-\tau)(m_2-m_1)>0$: $g$ is continuous, strictly decreasing, positive below $\essinf X$ and negative above $\esssup X$, so it has a unique root, the unique minimizer $e_\tau[X]$, characterized by the first-order condition
\begin{equation}
\label{eq:foc}
\tau\,\E[(X-e_\tau[X])_+]=(1-\tau)\,\E[(e_\tau[X]-X)_+].
\end{equation}
(i)~At $\tau=\tfrac12$, $g(m)=\tfrac12\E[X-m]$, root $\E[X]$.
(ii)~$X\le Y$ a.s.\ makes $g_X\le g_Y$ pointwise; evaluating at the root of $g_X$ and using strict decrease of $g_Y$ orders the roots.
(iii)~$g_{X+c}(m)=g_X(m-c)$.
(iv)~$g(\esssup X)\le0$ gives $e_\tau\le\esssup X$ (symmetrically for $\essinf$); $\partial g/\partial\tau=\E[(X-m)_+]+\E[(m-X)_+]\ge0$ gives monotonicity in $\tau$; and if $e_\tau[X]$ stayed below some $b<\esssup X$ as $\tau\to1$, then $\tau\E[(X-b)_+]\le(1-\tau)\cdot2\supnorm{X}\to0$ would contradict $\Prob(X>b)>0$ via \eqref{eq:foc}.
(v)~$|X-Y|\le c$ a.s.\ gives $X\le Y+c$; apply (ii)+(iii) both ways. For $e_1=\esssup$, (ii),(iii),(v) are immediate from the definition. See also \citet{newey1987asymmetric,BELLINI201441}. \hfill$\square$

\section{The telescoping identity \eqref{eq:telescope}}
\label{app:telescope}

Solve the definition of $\epsilon_k$ for the reward, $r(s_k,a_k)=Q^*(s_k,a_k)-\gamma\max_{a'}Q^*(s_{k+1},a')+\epsilon_k$; for $k\le n-2$ expand $\max_{a'}Q^*(s_{k+1},a')=Q^*(s_{k+1},a_{k+1})+\Delta^*(s_{k+1},a_{k+1})$, and for $k=n-1$ keep the max (it cancels the bootstrap term of \eqref{eq:Z}). Substituting into $\Zn{Q^*}$, the sum $\sum_{k=0}^{n-2}[\gamma^kQ^*(s_k,a_k)-\gamma^{k+1}Q^*(s_{k+1},a_{k+1})]+\gamma^{n-1}Q^*(s_{n-1},a_{n-1})$ telescopes to $Q^*(s,a)$, leaving \eqref{eq:telescope}. Under Assumption~\ref{ass:det}, $\epsilon_k\equiv0$. \hfill$\square$

\section{Proof of Theorem~\ref{thm:contraction}}
\label{app:contraction}

Fix $(s,a)$ and couple $\Zn{Q_1}(s,a)$, $\Zn{Q_2}(s,a)$ on the same trajectory (same actions, transitions, rewards). Rewards cancel and, since $|\max_{a'}Q_1(s',a')-\max_{a'}Q_2(s',a')|\le\supnorm{Q_1-Q_2}$ for every $s'$ (bound each $Q_1(s',a')$ by $Q_2(s',a')+\supnorm{Q_1-Q_2}$ and take maxima both ways), $|\Zn{Q_1}-\Zn{Q_2}|\le\gamma^n\supnorm{Q_1-Q_2}$ almost surely. Lemma~\ref{lem:expectile}(v) transfers this to the expectiles; take $\sup_{(s,a)}$. If $\supnorm{Q}\le Q_{\max}$ then $|\Zn{Q}|\le Q_{\max}$ pointwise, and Lemma~\ref{lem:expectile}(iv) keeps $\Top Q$ in the same ball, which is complete under $\supnorm{\cdot}$; the Banach fixed-point theorem \citep{banach1922operations,puterman2014markov} concludes. \hfill$\square$

\section{Proof of Theorems~\ref{thm:bias} and~\ref{thm:stoch}}
\label{app:bias}

\textbf{Moments of $W_n$.} Write $W_n=\sum_{j=0}^{n-1}\gamma^j\xi_j$ with $\xi_0\triangleq\epsilon_0$ and $\xi_j\triangleq\epsilon_j-\Delta^*(s_j,a_j)$ for $1\le j\le n-1$. Then $|\epsilon_j|\le R_{\max}+(1+\gamma)Q_{\max}$ and $0\le\Delta^*\le2Q_{\max}$, so $|\xi_j|\le M$. By the tower property, $\E_{s,a}[W_n]=-\mu_n(s,a)\le0$. Bounding every covariance by Cauchy--Schwarz, $\mathrm{Var}_{s,a}(W_n)\le M^2\big(\sum_{j=0}^{n-1}\gamma^j\big)^2\le\Vinf$; $M$ and $\Vinf$ do not depend on $n$. These are the constants of LQL's Appendix~C.1.

\textbf{Theorem~\ref{thm:bias}.} Under Assumption~\ref{ass:det}, \eqref{eq:telescope} reads $\Zn{Q^*}=Q^*(s,a)-\sum_{k=1}^{n-1}\gamma^k\Delta^*(s_k,a_k)$ pathwise.
(a)~$\mathcal T^n_{1/2}$ is the conditional mean (Lemma~\ref{lem:expectile}(i)); take expectations; each summand of $\mu_n$ is nonnegative, so $\mu_n$ is nondecreasing in $n$.
(b)~$\Top$ is monotone in its argument (couple trajectories; Lemma~\ref{lem:expectile}(ii)) and, at any fixed argument, monotone in $\tau$ (Lemma~\ref{lem:expectile}(iv)). By induction, iterates from a common initialization stay ordered, $(\mathcal T^n_\tau)^mQ_0\le(\mathcal T^n_{\tau'})^mQ_0$ for $\tau\le\tau'$; Theorem~\ref{thm:contraction} passes the order to the limits.
(c)~For every in-support $(s,a)$, Assumption~\ref{ass:coverage} gives
\[
\begin{aligned}
e_1[\Zn{Q^*}(s,a)]
&=
Q^*(s,a)
-\operatorname*{ess\,inf}_{\Prob_{s,a}}
  \sum_{k=1}^{n-1}\gamma^k\Delta^*(s_k,a_k) \\
&=Q^*(s,a).
\end{aligned}
\]
Hence $(\mathcal T_1^nQ^*)(s,a)=Q^*(s,a)$ on the covered in-support pairs.
(d)~$\Zn{Q^*}\le Q^*(s,a)$ pathwise; Lemma~\ref{lem:expectile}(ii)--(iii) against the constant give $\Top Q^*\le Q^*$. Since $\Top$ is monotone, the iterates $(\Top)^mQ^*$ are nonincreasing and converge (Theorem~\ref{thm:contraction}) to the unique fixed point, whence $\Qfix\le Q^*$. \hfill$\square$

\textbf{Corollary~\ref{cor:implicit}.} At $\tau=1$ the fixed-point equation reads $Q^{n}_{1}(s,a)=\esssup\big[G_{0:n}+\gamma^{n}\max_{a'}Q^{n}_{1}(s_n,a')\big]$ under $\Prob_{s,a}$, so the bracketed quantity is $\Prob_{s,a}$-a.s.\ at most $Q^{n}_{1}(s,a)$; on logged segments this is the case $j=1$. For $j>1$, apply the case $j=1$ at $(s_{t+n},a_{t+n})$, in-support for almost every logged segment, and chain with $\max_{a'}Q^{n}_{1}(s_{t+n},a')\ge Q^{n}_{1}(s_{t+n},a_{t+n})$; induction over $j$ concludes. \hfill$\square$

\textbf{Auxiliary lemma} \emph{(displacement identity)}\textbf{.} For $\tau\in(\tfrac12,1)$ and $m=e_\tau[X]$: setting $A=\E[(X-m)_+]$, $B=\E[(m-X)_+]$, the first-order condition \eqref{eq:foc} gives $B=\tfrac{\tau}{1-\tau}A$, and $A-B=\E[X]-m$ (identity $u_+-(-u)_+=u$); solving,
\begin{equation}
\label{eq:displacement}
m=\E[X]+\tfrac{2\tau-1}{1-\tau}\,\E[(X-m)_+].
\end{equation}

\textbf{Theorem~\ref{thm:stoch}.}
(a)~\emph{Lower:} monotonicity in $\tau$ and $e_{1/2}[W_n]=-\mu_n$. \emph{Upper:} in \eqref{eq:displacement} with $X=W_n$, use $m\ge\E[W_n]$ to dominate $(W_n-m)_+\le(W_n-\E[W_n])_+$, and $\E[(W_n-\E W_n)_+]=\tfrac12\E|W_n-\E W_n|\le\tfrac12\sigma_n$ (mean-zero positive-part identity, then Cauchy--Schwarz); so $m\le-\mu_n+\kt\sigma_n$.
(b)~If $\sigma_n=0$, then
$W_n=-\mu_n$ almost surely, so
$B_\tau=-\mu_n\le0$ for every $\tau$, and the stated threshold equals
$1$. If $\sigma_n>0$, part~(a) shows that
$\kt\sigma_n\le\mu_n$ is sufficient for $B_\tau\le0$.
Solving
\[
\frac{2\tau-1}{2(1-\tau)}\sigma_n\le\mu_n
\]
for $\tau$ yields
\[
\tau\le
\frac{\sigma_n+2\mu_n}{2\sigma_n+2\mu_n},
\]
which is the stated threshold.
(c)~$\supnorm{\Qfix-Q^*}\le\supnorm{\Top\Qfix-\Top Q^*}+\supnorm{\Top Q^*-Q^*}\le\gamma^n\supnorm{\Qfix-Q^*}+\sup_{s,a}|B_\tau|$; rearrange and bound $|B_\tau|\le\mu_n+\kt\sigma_n\le\sup_{s,a}\mu_n+\kt\sqrt{\Vinf}$ using (a) and the moment bounds. \hfill$\square$

\section{Algorithm}
\label{app:algorithm}

\begin{algorithm}[h]
\caption{ENQ critic update on a sampled segment}
\label{alg:enq}
\begin{algorithmic}[1]
\STATE \textbf{Input:} dataset $\mathcal{D}$; critics $\{Q_{\theta,i}\}_{i=1}^K$; targets $\{Q_{\bar\theta,i}\}$; actor $\pi_\phi$; horizon $n$; expectile $\tau$; penalty $\rho$; step size $\eta$; soft-update rate $\zeta$
\STATE Sample $(s_{t:t+n},a_{t:t+n-1},r_{t:t+n-1},d_{t:t+n-1},m_{t+1:t+n})\sim\mathcal{D}$
\STATE $G\leftarrow\sum_{k=0}^{n-1}\gamma^{k}r_{t+k}$;\ \ $v_t\leftarrow\prod_{k=t}^{t+n-2}(1-d_k)$
\STATE $a^*\sim\pi_\phi(\cdot\mid s_{t+n})$ for a stochastic actor (or $a^*=\pi_\phi(s_{t+n})$ for a deterministic actor)
\STATE $Y\leftarrow G+\gamma^{n}m_{t+n}\bar Q^{\rho}_{\bar\theta}(s_{t+n},a^*)$
\hfill(Eq.~\ref{eq:agg}--\ref{eq:enq_target})
\STATE $\delta_i\leftarrow \sg[Y]-Q_{\theta,i}(s_t,a_t)$,\ \ $i=1,\dots,K$
\STATE $\theta\leftarrow\theta-\eta\nabla_\theta\big[v_t\cdot\tfrac1K\sum_i\ell_\tau(\delta_i)\big]$
\STATE $\bar\theta\leftarrow\zeta\theta+(1-\zeta)\bar\theta$
\end{algorithmic}
\end{algorithm}

\section{Scope of the analysis}
\label{app:scope}
The analysis relies on three idealizations, each of which also appears in prior analyses of IQL and LQL \citep{kostrikov2021offline,abraham2026long}. \emph{(i)~Operator vs.\ algorithm:} the theorems concern the exact operator; the algorithm approximates one application by stochastic gradient descent (SGD) on \eqref{eq:enq_loss} with function approximation and a target network. \emph{(ii)~Bootstrap:} the analysis uses the greedy maximization; in practice a behavior-regularized actor serves as its proxy, and all contraction and monotonicity arguments hold for any fixed bootstrap policy. \emph{(iii)~Bootstrap-error amplification:} under function approximation, any bootstrapped scheme can amplify an error at a sparsely corrected state; this issue is distinct from the expectile-placement comparison but may interact with it under function approximation. Relatedly, finite-sample estimation of high expectiles can be sensitive to noisy upper-tail targets; this sensitivity may compound the familiar overestimation caused by maximizing over noisy action values \citep{thrun2014issues,hasselt2010double,vanhasselt2016deep}. This is one possible explanation for the failures at extreme $\tau$ and long horizons in ablation~(A1).

%
%
\newcommand{\ProfileHorizon}{8}

\section{Experimental Details}
\label{app:expdetails}

\subsection{Benchmarks, datasets, and task selection}
\label{app:tasks}

We evaluate on OGBench~\citep{park2025ogbench} and
RoboMimic~\citep{robomimic2021}. Our OGBench evaluation covers five
task groups: \texttt{cube-double}, \texttt{cube-triple},
\texttt{humanoidmaze-medium}, \texttt{antmaze-giant}, and
\texttt{humanoidmaze-giant}. For each group, we evaluate all five
single-task variants, \texttt{task1} through \texttt{task5}, yielding
25 OGBench task instances. We additionally evaluate the
\texttt{can} and \texttt{square} RoboMimic tasks, for a total of 27
task instances.

\begin{figure*}[t]
  \centering
  \begin{tabular}{cccc}
    \includegraphics[width=0.21\linewidth]{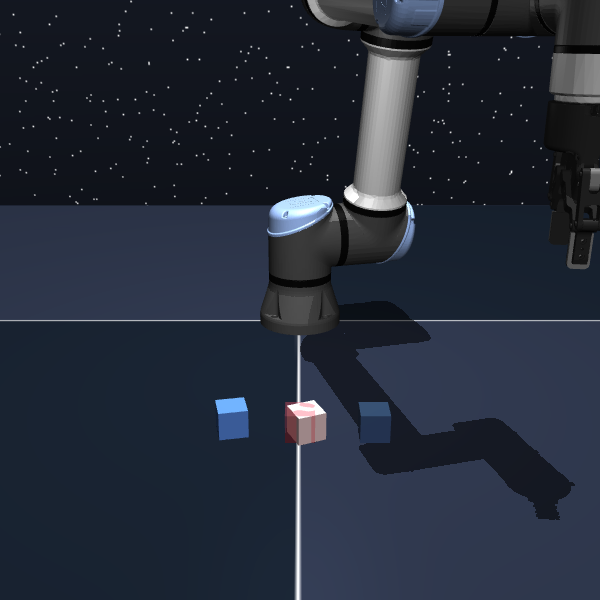} &
    \includegraphics[width=0.21\linewidth]{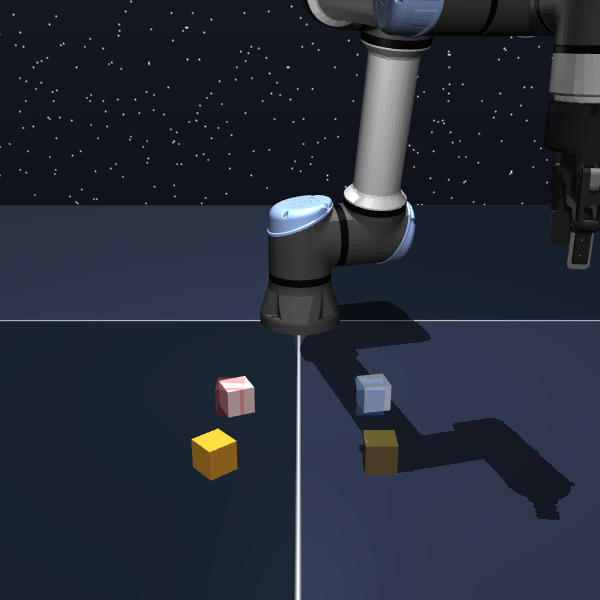} &
    \includegraphics[width=0.21\linewidth]{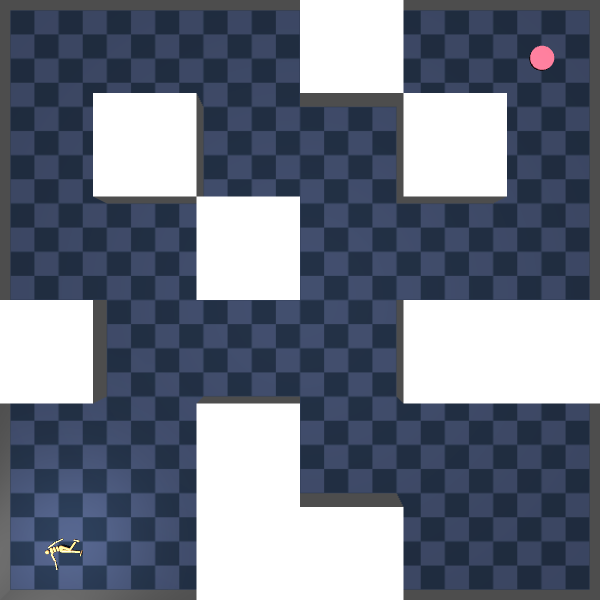} &
    \includegraphics[width=0.21\linewidth]{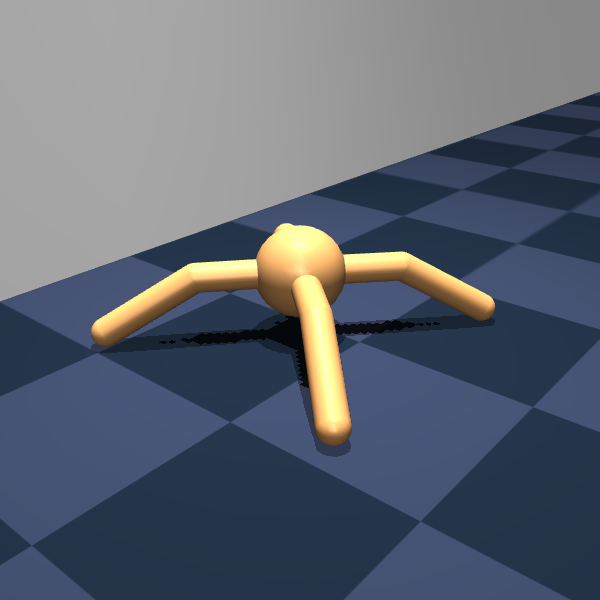} \\
    {\small (a) \texttt{cube-double}} &
    {\small (b) \texttt{cube-triple}} &
    {\small (c) \texttt{humanoidmaze-medium}} &
    {\small (d) \texttt{antmaze-giant}} \\[4pt]
    \includegraphics[width=0.21\linewidth]{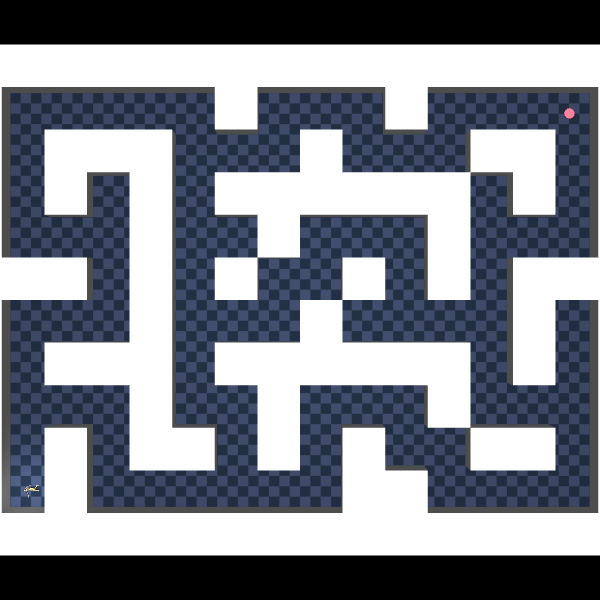} &
    \includegraphics[width=0.21\linewidth]{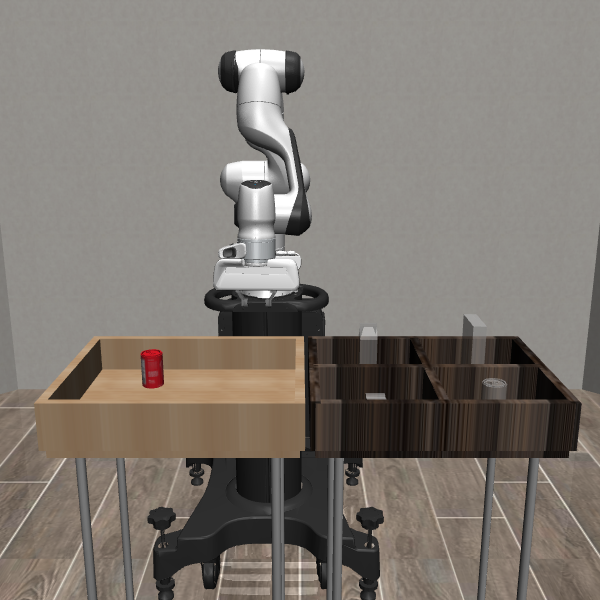} &
    \includegraphics[width=0.21\linewidth]{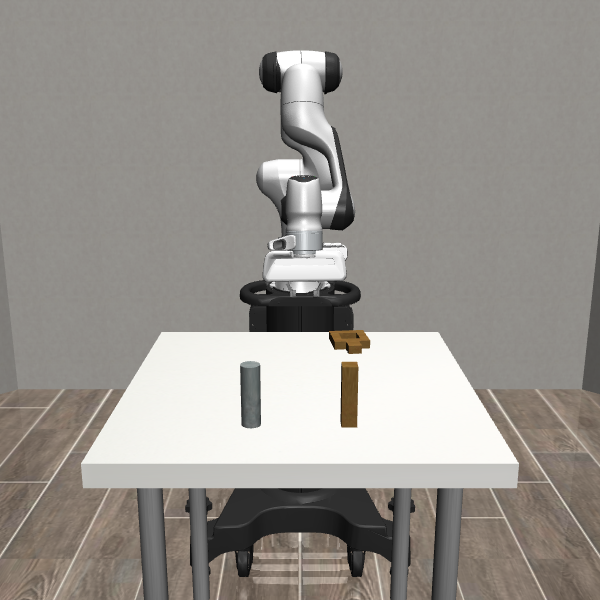} &
    {} \\
    {\small (e) \texttt{humanoidmaze-giant}} &
    {\small (f) \texttt{can}} &
    {\small (g) \texttt{square}} &
    {}
  \end{tabular}
\caption{\textbf{Evaluated OGBench and RoboMimic domains.}
Each OGBench group contains five single-task variants;
\texttt{can} and \texttt{square} use the RoboMimic multi-human
datasets. The \texttt{antmaze-giant} and
\texttt{humanoidmaze-giant} environments share the giant-maze
layout but use different agents.}
  \label{fig:task_panel}
\end{figure*}

The OGBench single-task variants fix an evaluation goal and relabel
the corresponding exploratory dataset with the benchmark reward. The
cube domains use play-style datasets, whereas the maze domains use
navigate-style datasets. For \texttt{humanoidmaze-giant}, we use the
100M-transition navigate-style dataset released by the OGBench
authors and also used by LQL~\citep{abraham2026long}. For RoboMimic,
we use the multi-human datasets containing demonstrations from
operators of varying proficiency. Table~\ref{tab:dataset_metadata}
summarizes the domain metadata, and Figure~\ref{fig:task_panel}
visualizes the evaluated domains.

\begin{table*}[t]
  \centering
  \caption{\textbf{Dataset and environment metadata.} Each OGBench
  row contains five single-task variants. Episode length is the
  environment time limit and action dimension is the dimension of one
  environment action.}
  \label{tab:dataset_metadata}
  \begin{small}
  \setlength{\tabcolsep}{5pt}
  \begin{tabular}{lrrr}
    \toprule
    Task group & Dataset size & Episode length & Action dimension \\
    \midrule
    \texttt{cube-double}          & 1M     & 500  & 5  \\
    \texttt{cube-triple}          & 3M     & 1000 & 5  \\
    \texttt{humanoidmaze-medium}  & 2M     & 2000 & 21 \\
    \texttt{antmaze-giant}        & 1M     & 1000 & 8  \\
    \texttt{humanoidmaze-giant}   & 100M   & 4000 & 21 \\
    \texttt{can}                  & 62,756 & 500  & 7  \\
    \texttt{square}               & 80,731 & 500  & 7  \\
    \bottomrule
  \end{tabular}
  \end{small}
\end{table*}

We use the benchmark success indicator and report the percentage of
successful evaluation episodes. The labels
\texttt{humanoid-md} and \texttt{humanoidmaze-medium} refer to the
same task group throughout the paper and supplement.

\subsection{Policy extraction and baseline provenance}
\label{app:policy_extraction}

We focus on flow-matching policy classes. On OGBench and RoboMimic,
we evaluate Flow Q-Learning (FQL)~\citep{park2025flow} and its
action-chunked variant~\citep{li2026reinforcement}. The action-chunked
ENQ actor and critic operate on chunks of \(h=4\) consecutive actions.
On \texttt{humanoidmaze-giant}, we use the Best-of-\(N\) policy
extraction used by LQL for this domain, with \(N=16\) candidates.
The flow ODE is integrated with 10 Euler steps.

The TD, TD-\(n\), and LQL entries in the main comparative tables are
taken directly from the results reported by
\citet{abraham2026long}; they were not rerun for the present paper.
Consequently, those cells retain the original LQL implementation and
hyperparameters, including its action-chunk size \(h=5\), its
domain-specific policy coefficients, ensemble-mean target aggregation
on OGBench, and ensemble-minimum aggregation on RoboMimic. ReBRAC and
IQL values are likewise taken from the cited prior work indicated in
the main-paper table captions. All additional ENQ ablations, the
ten-critic LQL scaling experiment, and the stochastic-transition
controls were run in our JAX codebase derived from the released LQL
implementation.

\subsection{Offline-to-online training protocol}
\label{app:offline_online_protocol}

Each ENQ run consists of \(10^6\) offline gradient updates followed by
\(10^6\) online environment steps. During the online phase, the replay buffer is initialized from the
offline data and one newly collected transition is inserted per
environment step. After an initial 5,000-step online collection
period, one gradient update is performed per environment step
(UTD ratio $1$). The replay
capacity is \(2\times10^6\) transitions for the standard datasets;
the implementation streams or replaces offline shards for datasets
that exceed the in-memory capacity. Offline data remain available
during online fine-tuning through the initialized replay buffer.

For action-chunked policies, the actor generates one chunk and the
environment executes its actions sequentially. A transition is stored
after every executed action. Following the initial 5,000-step online
collection period, one gradient update is performed per executed
environment action. Terminal bootstraps are masked, and sampled segments crossing
an episode boundary are discarded by the validity mask defined in the
main paper.

Across ENQ--LQL comparisons, we match the number of sampled
transitions processed per gradient update:
\[
B_{\mathrm{ENQ}}n=B_{\mathrm{LQL}}L.
\]
In the default setting, ENQ uses \(B_{\mathrm{ENQ}}=256\) and \(n=4\),
while the published LQL baseline uses 128 trajectories of length
\(L=8\); both process 1024 transitions per update. On
\texttt{humanoidmaze-giant}, ENQ uses \(B_{\mathrm{ENQ}}=2048\) and
\(n=4\), while LQL uses 128 trajectories of length \(L=64\); both
process 8192 transitions per update.

\subsection{Common hyperparameters}
\label{app:common_hyperparams}

Table~\ref{tab:common_hyperparams} reports the common configuration.
The ENQ implementation inherits the architecture and optimization
defaults of the LQL codebase. The expectile level and backup horizon
are fixed to \(\tau=0.8\) and \(n=4\) for every main ENQ task, without
task-specific tuning. For the published \texttt{humanoidmaze-giant} comparison, the
TD-$n$ baseline uses $n=4$.

\begin{table}[t]
  \centering
  \caption{\textbf{Common hyperparameters.} ``Published baselines''
  refers to the TD, TD-\(n\), and LQL values copied from
  \citet{abraham2026long}. Domain-specific exceptions are listed in
  Table~\ref{tab:task_hyperparams}.}
  \label{tab:common_hyperparams}
  \begin{small}
  \setlength{\tabcolsep}{4pt}
  \begin{tabular}{p{0.39\linewidth}p{0.25\linewidth}p{0.25\linewidth}}
    \toprule
    Parameter & ENQ runs & Published baselines \\
    \midrule
    Optimizer & Adam & Adam \\
    Learning rate & \(3\times10^{-4}\) & \(3\times10^{-4}\) \\
    Target soft-update rate & \(5\times10^{-3}\) & \(5\times10^{-3}\) \\
    UTD ratio & 1 & 1 \\
    Hidden layers & \(4\times512\) & \(4\times512\) \\
    Activation & GELU & GELU \\
    Critic LayerNorm & Yes & Yes \\
    Actor LayerNorm & No & No \\
    Flow integration steps & 10 & 10 \\
    Main critic ensemble size & \(K=2\) & \(K=2\) \\
    Offline updates & \(10^6\) & \(10^6\) \\
    Online environment steps & \(10^6\) & \(10^6\) \\
    Replay capacity & \(2\times10^6\) & \(2\times10^6\) \\
    ENQ expectile & \(\tau=0.8\) & Not applicable \\
    ENQ backup horizon & \(n=4\) & Not applicable \\
TD-$n$ horizon
& Not applicable
& $n=8$ (default) \\
    LQL trajectory length & Not applicable & \(L=8\) \\
    Action-chunk size & \(h=4\) & \(h=5\) \\
    Best-of-\(N\) candidates & \(N=16\) & \(N=16\) \\
    Evaluation interval & \(10^5\) steps & As reported by LQL \\
    Evaluation episodes & 50 & As reported by LQL \\
    Seeds per task & 4 & 4 \\
    \bottomrule
  \end{tabular}
  \end{small}
\end{table}

\subsection{Task-specific hyperparameters}
\label{app:task_hyperparams}

We use \(\gamma=0.99\) for manipulation tasks and
\(\gamma=0.995\) for locomotion and navigation tasks, following the
domain-dependent convention used in recent generative-policy
value-learning work~\citep{li2026qlearningadjointmatching}. The FQL
behavior-regularization coefficients follow LQL except on
\texttt{antmaze-giant}, where we use \(\alpha=3\), as in QAM. The
same \(\alpha\) is used for FQL and action-chunked FQL.

For ENQ, the target-ensemble aggregation is
\[
\bar Q^\rho(s,a)
=
\frac1K\sum_{i=1}^{K}Q_i(s,a)
-\rho\,\operatorname{std}_{i}Q_i(s,a).
\]
Following QAM, we set \(\rho=0\) on both humanoidmaze domains and
\(\rho=0.5\) elsewhere. Thus, the method-specific ENQ settings
\(\tau=0.8\) and \(n=4\) remain fixed across all tasks; only
environment and policy-extraction settings inherited from the
benchmark recipes vary by domain.

\begin{table*}[t]
  \centering
  \caption{\textbf{Task-specific hyperparameters and provenance.}
  \(B\) is the ENQ segment batch size. The final column records the
  corresponding settings used by the published LQL/TD baselines.}
  \label{tab:task_hyperparams}
  \begin{small}
  \resizebox{\textwidth}{!}{%
  \begin{tabular}{llllrrrl}
    \toprule
    Task group & ENQ policy & \(\gamma\) & \(\alpha\) or \(N\) &
    \(\rho\) & \(B\) & Dataset & Published LQL/TD setting \\
    \midrule
    \texttt{cube-double}
      & FQL / chunked FQL & 0.99 & \(\alpha=100\) & 0.5 & 256
      & play, 1M
      & \(\gamma=.99,\alpha=100,h=5,\) mean \\
    \texttt{cube-triple}
      & FQL / chunked FQL & 0.99 & \(\alpha=100\) & 0.5 & 256
      & play, 3M
      & \(\gamma=.99,\alpha=100,h=5,\) mean \\
    \texttt{humanoidmaze-medium}
      & FQL / chunked FQL & 0.995 & \(\alpha=10\) & 0 & 256
      & navigate, 2M
      & \(\gamma=.99,\alpha=10,h=5,\) mean \\
    \texttt{antmaze-giant}
      & FQL / chunked FQL & 0.995 & \(\alpha=3\) & 0.5 & 256
      & navigate, 1M
      & \(\gamma=.99,\alpha=5,h=5,\) mean \\
    \texttt{humanoidmaze-giant}
      & Best-of-\(N\) & 0.995 & \(N=16\) & 0 & 2048
      & navigate, 100M
     & $\gamma=.995,N=16,\text{TD-}4,L=64,$ mean \\
    \texttt{can}, \texttt{square}
      & FQL / chunked FQL & 0.99 & \(\alpha=250\) & 0.5 & 256
      & multi-human
      & \(\gamma=.99,\alpha=250,h=5,\) minimum \\
    \bottomrule
  \end{tabular}%
  }
  \end{small}
\end{table*}

\subsection{Evaluation and statistical reporting}
\label{app:evaluation}

We run four seeds per task and evaluate every \(100{,}000\) training
steps using 50 evaluation episodes. Final performance is the
evaluation at the end of the \(10^6\)-step online fine-tuning phase.
Unless a figure caption states otherwise, uncertainty bands and
brackets are 95\% confidence intervals obtained from 1000
curve-level bootstrap resamples. A complete seed-level training curve
is resampled as one unit. For multi-task aggregates, runs are
resampled independently within each task, the mean is computed within
each task, and the resulting task means are combined with equal
weight. Figures whose captions report one standard deviation instead
use the empirical standard deviation across the four seeds.

\subsection{A1: expectile and horizon sweep}
\label{app:a1}

We evaluate the interaction between the expectile and backup horizon
on \texttt{cube-triple-task2} and
\texttt{antmaze-giant-task1}, using both FQL and action-chunked FQL.
These are the default manipulation and locomotion tasks used in the
OGBench evaluation convention adopted by
QAM~\citep{li2026qlearningadjointmatching}. We sweep
\[
\tau\in\{0.5,0.8,0.9,0.99\},
\qquad
n\in\{4,8,16\}.
\]
For every horizon, we keep the number of sampled transitions per
update fixed at 1024:
\[
(B,n)\in\{(256,4),(128,8),(64,16)\}.
\]
All other settings, including the policy-extraction mechanism,
network architecture, optimizer, ensemble size \(K=2\), and
domain-specific conservative coefficient, are held fixed. The
\(\tau=0.5\) configurations are the matched symmetric TD-\(n\)
controls.

\subsection{A2: critic-ensemble scaling}
\label{app:a2}

\paragraph{Performance scaling.}
We study performance at \(K=2\) and \(K=10\) on all five
\texttt{cube-triple} tasks with action-chunked policy extraction. We
choose this group because the \(K=2\) ENQ and LQL aggregate results are
nearly matched in the main comparison. The \(K=2\) values are the
main-experiment configurations. For \(K=10\), only the number of
critics is changed: ENQ retains its \(\tau=0.8\), \(n=4\),
\(h=4\), \(\rho=0.5\), and \(B=256\) configuration, while LQL retains
the published \texttt{cube-triple} action-chunking configuration and
changes its ensemble from two to ten critics. Both methods use four
seeds per task.

\paragraph{Throughput profiling.}
We profile ENQ and LQL on
\texttt{antmaze-giant-navigate-singletask-task2-v0} with FQL policy
extraction, batch size \(B=128\), matched horizon
\(n=L=\ProfileHorizon\), \(\alpha=5\), \(\gamma=0.99\), and
ensemble-mean aggregation (\(\rho=0\)). We evaluate
\(K\in\{2,5,10,20,50\}\) using seed 10001.

Each configuration executes 10,000 offline training iterations with
no online interaction, evaluation, or checkpointing. The first 9000
iterations serve as compilation and warm-up. Throughput is measured
over the final 1000 iterations.
\begin{table}[t]
  \centering
  \caption{\textbf{Measured profiling throughput.} Ratio is
  ENQ iterations per second divided by LQL iterations per second.}
  \label{tab:profiling_raw}
  \begin{small}
  \begin{tabular}{rrrr}
    \toprule
    \(K\) & LQL iter./s & ENQ iter./s & ENQ/LQL \\
    \midrule
     2 & 662.3 & 838.2 & \(1.27\times\) \\
     5 & 517.2 & 718.0 & \(1.39\times\) \\
    10 & 385.3 & 561.6 & \(1.46\times\) \\
    20 & 249.8 & 387.4 & \(1.55\times\) \\
    50 & 105.5 & 187.1 & \(1.77\times\) \\
    \bottomrule
  \end{tabular}
  \end{small}
\end{table}

\subsection{A3: expectile placement}
\label{app:a3}

To isolate the effect of placing the asymmetric regression directly
on the action-value residual, we implement an \(n\)-step IQL-style
baseline with an auxiliary state-value network. Let
\[
\widehat Y^{V}_{t,n}
=
G_{t:t+n}
+
\gamma^n m_{t+n}V_{\bar\psi}(s_{t+n}),
\]
where \(V_{\bar\psi}\) is a target value network. The online value
network is trained by
\[
\mathcal L_V(\psi)
=
\mathbb E_{\mathcal D}\!\left[
v_t\,
\ell_\tau\!\left(
\widehat Y^{V}_{t,n}-V_\psi(s_t)
\right)
\right],
\]
and each Q-critic is trained symmetrically toward the same target:
\[
\mathcal L_Q(\theta)
=
\mathbb E_{\mathcal D}\!\left[
v_t\,
\frac1K\sum_{i=1}^{K}
\left(
\widehat Y^{V}_{t,n}-Q_{\theta,i}(s_t,a_t)
\right)^2
\right].
\]
The total critic objective is
\(\mathcal L_Q+\mathcal L_V\). The value network is a single
four-layer, width-512 network with the same LayerNorm convention as
the critic. Its target copy is initialized from the online value
network and soft-updated with rate \(5\times10^{-3}\). No critic
aggregation or \(\rho\) penalty enters the bootstrap because the
target uses the scalar value \(V_{\bar\psi}(s_{t+n})\).

We compare this baseline with ENQ on
\texttt{antmaze-giant-task1} and \texttt{cube-triple-task2}. We
transfer the complete corresponding ENQ configuration without
retuning: \(\tau=0.8\), \(n=4\), action-chunk size \(h=4\),
\(K=2\), the same actor and policy coefficient, the same optimizer and
network dimensions, the same batch size, and the same
\(10^6\)-offline/\(10^6\)-online training protocol.

\subsection{A4: stochastic-transition experiments}
\label{app:a4}

We construct stochastic versions of
\texttt{cube-triple-task2} by perturbing each action independently
before environment execution:
\[
\widetilde a_t
=
\operatorname{clip}\!\left(
a_t+\epsilon_t,-1,1
\right),
\qquad
\epsilon_t\sim\mathcal N(0,\sigma^2I).
\]
For each noise level, we recollect a separate offline dataset using
the OGBench oracle and the benchmark data-collection scripts. The
intended oracle action \(a_t\) is stored in the dataset, while the
perturbed action \(\widetilde a_t\) is executed and therefore
determines the observed transition. During online fine-tuning, the learned policy action is perturbed in
the same way before execution. The same perturbation level is also
applied during evaluation, so each reported success rate measures
performance in the corresponding stochastic environment rather than
under clean evaluation. Mild action perturbations can occasionally
improve performance relative to the unperturbed setting by increasing
state--action coverage, an effect also observed in OGBench-style data
collection.

For the noise-level comparison, we use
\[
\sigma\in\{0,0.05,0.25\}.
\]
ENQ, TD-\(n\), and LQL use ten critics, ensemble-mean target
aggregation, and \(\rho=0\); ENQ uses \(\tau=0.8\). Every method is
trained for \(10^6\) offline updates followed by \(10^6\) online
environment steps.

At the strongest noise level, \(\sigma=0.25\), we sweep ENQ over
\[
\tau\in\{0.5,0.8,0.95,0.99\},
\qquad
\rho\in\{0,0.5\}.
\]
The same panel additionally reports two LQL controls, using
\(\rho=0\) and \(\rho=0.5\), respectively. All of these
configurations use \(K=10\), and all remaining optimization,
architecture, data, and training-budget settings are held fixed.
For \(\rho=0.5\), both methods use the mean-minus-\(0.5\)-standard-
deviation target aggregation.

\subsection{Hardware and software}
\label{app:hardware}

The implementation is written in JAX and builds on the released LQL
codebase. All experiments run for this work were performed on a
workstation equipped with an NVIDIA GeForce RTX~5090 GPU with
32\,GB of memory. The software environment used Python~3.11.15,
JAX and jaxlib~0.9.0.1, Flax~0.12.3, and Optax~0.2.7.

\end{document}